\documentclass{article} %
\usepackage[final]{colm2026_conference} %

\usepackage{microtype}
\usepackage{hyperref}
\usepackage{url}
\usepackage{booktabs}

\usepackage{lineno}

\usepackage{tikz}
\usepackage[dvipsnames]{xcolor}
\usepackage{graphicx} %
\hypersetup{
    colorlinks=true,
    linkcolor=blue,
    filecolor=magenta,      
    urlcolor=cyan
    }
\usepackage{amsmath}
\usepackage{booktabs}
\usetikzlibrary{trees}
\usepackage{xspace}
\usepackage{amsthm}  %
\usepackage[normalem]{ulem}  %
\usepackage{cleveref}
\usepackage[most]{tcolorbox}
\usepackage{wrapfig}
\usepackage{array}
\usepackage{tikz}
\usetikzlibrary{positioning}
\usepackage{subcaption}
\usepackage{multirow}
\usepackage{tabularx}
\usepackage{capt-of}

\usepackage{listings}
\usepackage{xcolor}
\usepackage[title]{appendix}

\lstdefinelanguage{json}{
  basicstyle=\ttfamily\small,
  upquote=false,
  showstringspaces=false,
  breaklines=true,
  frame=single,
  backgroundcolor=\color{gray!5},
  stringstyle=\color{blue},
  keywordstyle=\color{purple},
}

\newcommand{\xhdr}[1]{{\vspace{1mm}\noindent{\textbf{#1}}}}

\newcommand{\gptfour}[0]{{\tt GPT-4o}\xspace}  
\newcommand{\gptfive}[0]{{\tt GPT-5}\xspace}  
\newcommand{\geminithree}[0]{{\tt Gemini 3}\xspace}  
\newcommand{\geminitwo}[0]{{\tt Gemini 2.5}\xspace}  
\newcommand{\qwentwo}[0]{{\tt Qwen 2.5}\xspace}  
\newcommand{\qwenthree}[0]{{\tt Qwen 3}\xspace}  
\newcommand{\llamathree}[0]{{\tt Llama 3.1}\xspace}
\newcommand{\llamafour}[0]{{\tt Llama 4}\xspace}

\newcommand{\mistral}[0]{{\tt Mistral}\xspace}
\newcommand{\ministral}[0]{{\tt Ministral 3}\xspace}

\newcommand{\eg}[0]{\textit{e.g.,}\xspace}
\newcommand{\ie}[0]{\textit{i.e.,}\xspace}
\newcommand{\vs}[0]{\textit{vs.}\xspace}

\theoremstyle{definition}
\newtheorem{mydef}{Definition}

\definecolor{darkblue}{rgb}{0, 0, 0.5}
\hypersetup{colorlinks=true, citecolor=darkblue, linkcolor=darkblue, urlcolor=darkblue}

\title{On Epistemic Diversity in Large Language Models}

\author{
\textbf{Elisabeth Kirsten\textsuperscript{1,3}},
\textbf{Nicole Krämer\textsuperscript{1,2}},
 \textbf{Muhammad Bilal Zafar\textsuperscript{1,3}}
\\
 \textsuperscript{1}UAR Research Center for Trustworthy Data Science and Security,
 \\
 \textsuperscript{2}University of Duisburg-Essen,
 \\
 \textsuperscript{3}Ruhr University Bochum
\\
 \small{
   \textbf{Correspondence:}
   \href{mailto:elisabeth.kirsten@rub.de}{\color{blue}elisabeth.kirsten@rub.de}
 }
}

\begin{document}

\ifcolmsubmission
\linenumbers
\fi

\maketitle

\begin{abstract}
Large language models (LLMs) are increasingly used not only to retrieve information, but to answer questions, explain, teach, and support inquiry. 
In such settings, evaluation cannot be exhausted by accuracy or alignment alone. 
A system may give a correct answer while still narrowing users' access %
to alternative valid answers, explanations, or reasoning routes. 
Drawing on the broader notion of \textbf{epistemic diversity} in philosophy and social epistemology, we formalize it in the context of LLMs as the range of valid answers, explanations, and reasoning routes that an LLM exposes to users.
We argue that epistemic diversity is
a useful evaluation dimension for settings where LLMs are used to support knowledge-intensive tasks.
We propose a preliminary framework for conceptualizing and measuring epistemic diversity in LLMs, and operationalize it in two domains.
We find that \textbf{frontier LLMs often exhibit epistemic narrowness}, repeatedly collapsing large valid answer spaces onto small canonical subsets.
These findings suggest that LLM evaluation should move beyond accuracy-oriented paradigms and treat epistemic diversity as an important dimension of model capability.
\end{abstract}

\section{Introduction}
Large language models (LLMs) are increasingly used not only as \textit{predictive tools}, but as \textit{knowledge tools} for answering questions, explaining concepts, drafting arguments, and supporting inquiry \citep{chatterji_chatgpt_usage}. 
In these settings, what matters is not only whether a model can produce a correct or acceptable answer. 
What also matters is the kinds of answers, explanations, and reasoning the model makes available to users. 
A system may be accurate and yet still be epistemically narrow if, in contexts where alternatives would be useful, it repeatedly presents only a few canonical routes despite the existence of multiple valid answers.

Much of the existing literature on diversity in AI asks whether different demographic, cultural, or political groups are represented or treated fairly \citep{hardt_equal_op, barocas-hardt-narayanan, guo_ceat, wangLargeLanguageModels2024}. 
Those are important concerns. 
But for systems increasingly used to mediate access to knowledge, an additional question arises: not only \textit{who is represented} in model outputs, but \textit{what knowledge and ways of knowing} are made available.

\begin{figure*}[t]
    \centering
    \includegraphics[width=0.9\linewidth]{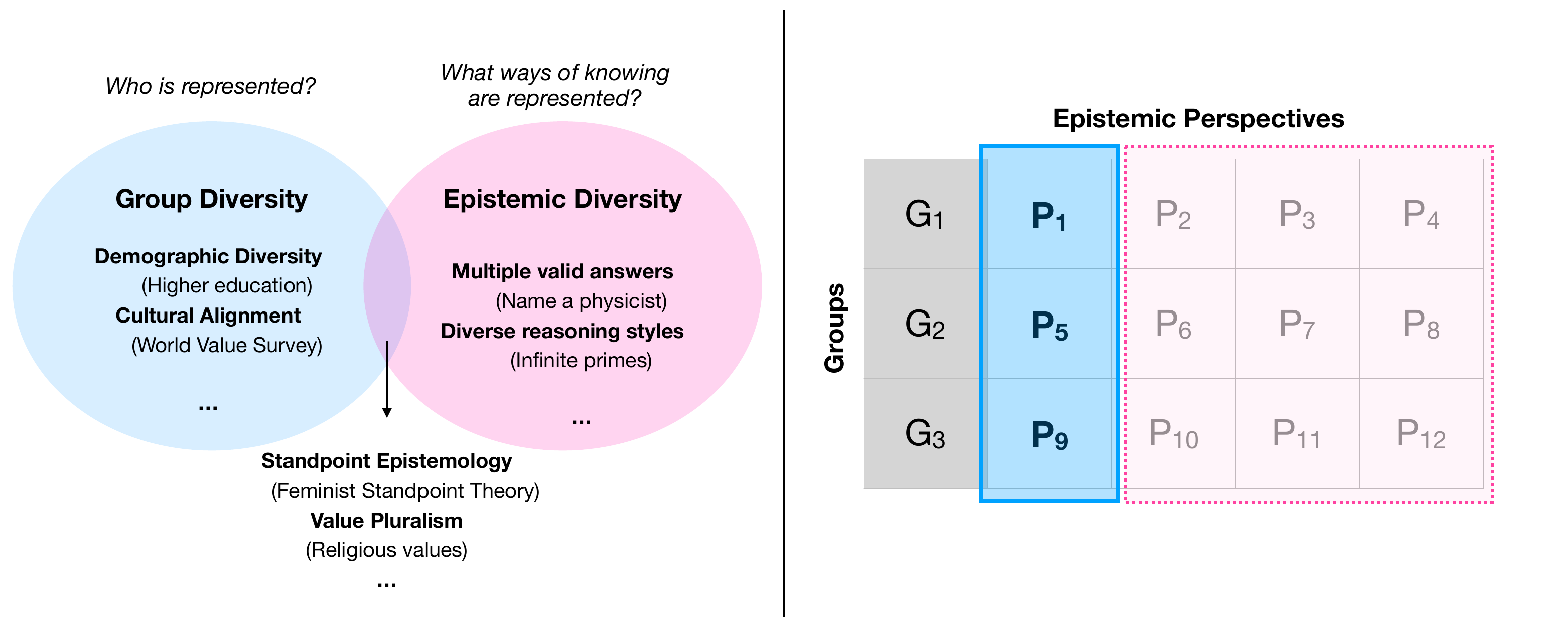}
    \caption{\textit{Left:} Epistemic and group diversity are distinct but may overlap. 
    \textit{Right:} Group diversity does not imply epistemic diversity: a system may represent all groups equally while surfacing only a narrow subset of the knowledge available across groups.
    }
    \label{fig:epistemic-vs-group}
\end{figure*}

To capture this complementary dimension,
we draw on the notion of \textbf{epistemic diversity}. By epistemic diversity, we mean the range of valid answers, explanations, or reasoning strategies that an LLM can produce for a given query. 
This notion is related to, but not reducible to, group-based diversity. 
Group diversity concerns representation across persons, communities, or viewpoints, whereas epistemic diversity concerns representation across the space of valid answers.
The two can overlap, for example when different social standpoints sustain different bodies of knowledge. 
But they can also be orthogonal. A system may represent multiple groups while still surfacing only a narrow subset of the knowledge available across them. 
Conversely, a system may expose multiple valid answers in a setting where the relevant issue is not representation but explanatory or conceptual breadth. 
The point is \textit{not that group diversity is unimportant, but that it does not exhaust the diversity-related desiderata relevant to LLMs}. See Figure~\ref{fig:epistemic-vs-group} for an explanation of the differences.

The need for this distinction becomes clear in ordinary LLM use. Consider a prompt such as ``Write about a famous physicist.''
There is no single uniquely correct answer: Many individuals would count as valid.  Likewise, when asked to prove that there are infinitely many primes, a model can give a correct proof while still being epistemically narrow if it always returns the same proof strategy despite the existence of many accepted alternatives. In such cases, the problem is not error in the familiar sense of accuracy. Rather, a large valid answer space is collapsed into a small canonical subset. This matters because users often seek not merely an output, but \textit{understanding}. A student may benefit from seeing multiple proof strategies, a reader from encountering figures beyond the most canonical examples, and a researcher from considering alternative framings of an underspecified question.
In all of these cases, accuracy remains indispensable, but diversity of valid answers matters too.

This concern regarding diversity of knowledge has deep roots in philosophy of science and social epistemology. Across these disciplines, diversity is valued not merely as a matter of inclusion, but as a means of better inquiry. Social epistemologists have argued that heterogeneous communities can reveal blind spots and correct distortions that no isolated reasoner can easily detect \citep{longinoScienceSocialKnowledge1990}. Pluralist traditions in philosophy of science have emphasized that inquiry often advances through the coexistence of multiple models, explanations, or methods rather than through immediate convergence on a single privileged representation \citep{kellertIntroductionPluralistStance2006, feyerabendMethod2002}. Standpoint epistemology likewise stresses that knowledge is situated, and that different positions can disclose different dimensions of a domain \citep{hardingWhoseScienceWhose, collinsBlackFeministThought2002}. Taken together, these traditions suggest that preserving multiple valid ways of knowing can itself be epistemically valuable, especially in settings oriented toward explanation, learning, and discovery.

Understanding often depends not only on receiving one correct answer, but on grasping alternatives, contrasts, and relations among possible explanations or solution paths.
A system that repeatedly presents only one canonical framing may therefore narrow the contrast space through which users achieve understanding, even when that framing is accurate.
This does not mean that models should always maximize diversity.
In many settings, a standard or canonical answer is useful.
The concern arises when alternatives would be valuable for learning, exploration, or pluralism, when users explicitly ask for different answers, or when repeated use across many users systematically concentrates attention on the small set of examples, explanations, or methods.

To study this phenomenon in LLMs, \textbf{we develop a framework that treats epistemic diversity not as arbitrary output variation, but as \uline{coverage over a space of valid answers}}. This framing helps distinguish epistemically meaningful diversity from mere stylistic variation, stochastic noise, or hallucination. It also clarifies when diversity should be expected. Some prompts are underspecified and admit multiple valid answers because important constraints are left open. Others are well specified but still admit many valid responses because the answer space is large or effectively unbounded, as in creative generation or problems with multiple proof strategies. 

Building on this framework, we evaluate epistemic diversity under multiple interaction protocols.
These protocols capture different ways in which users may encounter or seek multiple answers.
\textit{Accessible diversity} measures the diversity surfaced under ordinary prompting, including repeated exposure to the same prompt across users.
For example, if a user asks for a mathematical proof and the model ordinarily returns only one, then the model's accessible diversity is narrow, even though many other valid proof strategies exist.
We also include controlled probing protocols that approximate an upper bound on recoverable diversity, as well as more naturalistic protocols in which diversity-seeking intent is made explicit, such as asking for several answers at once or requesting alternatives in follow-up turns.
Such protocols may elicit induction-based, analytic, or combinatorial approaches that are rarely surfaced by default. 
Together, these protocols distinguish between diversity that is readily accessible in everyday use and diversity that is available only through more deliberate exploration.

We illustrate this framework in two settings that instantiate different answer space structures. Across both, frontier LLMs often exhibit what we call \textbf{epistemic narrowness}: They repeatedly concentrate on a small subset of valid outputs even when substantially more diversity is available. While stronger prompting can recover some additional diversity, a substantial gap remains.

\textbf{Our central claim is that epistemic diversity matters because LLMs increasingly function as knowledge sources: They do not merely produce answers, but shape which explanations, examples, and reasoning paths become salient and available to users.}

We make four contributions.
First, we distinguish epistemic diversity from group-based diversity in the context of LLM evaluation, focusing on the range of valid answers, examples, explanations, and reasoning strategies a model makes available. 
Second, we develop a measurement framework that operationalizes epistemic diversity as coverage over valid answer spaces.
Third, we show that measured diversity depends on how a model is queried, and define interaction protocols that separate the diversity surfaced by default from the diversity recoverable under controlled probing and under explicit diversity-seeking requests.
Fourth, we apply the framework in two controlled domains and show that current frontier models often exhibit epistemic narrowness.
We begin in Section 2 by reviewing how diversity is currently conceptualized in AI research,
and motivate epistemic diversity as a complementary perspective.

\section{From Group Diversity to Epistemic Diversity}
\label{sec:background_diversity}

Diversity has emerged as a central desideratum in AI and ML.
Multiple justifications have been provided for ensuring diversity. For instance, diversity is often invoked as a property that is ethically imperative, normatively desirable, and leads to optimal outcomes from a utilitarian perspective~\citep{barocas-hardt-narayanan,sep-discrimination,sep-equal-opportunity,Hooker_2014}.
A large body of work has focused on operationalizing and improving diversity in AI/ML systems~\citep{buolamwiniGenderShadesIntersectional2018,mitchellModelCardsModel2019,guoBenchmarkingLinguisticDiversity2025,zafar_disparate_mistreatment,hardt_equal_op}.
However, \textit{diversity} encompasses a wide range of meanings, and current approaches largely focus on \textit{who is represented} in model behavior. 
In this section, we review these strands of work and identify the gap that motivates our notion of \textit{epistemic diversity}.

\subsection{Current Conceptualization of AI Diversity}
\label{sec:group_diversity}
Current discourse on diversity in AI can be divided into several overlapping categories.

\xhdr{1. Demographic diversity.}
This branch aims to ensure that different demographic groups are adequately represented in data and model outcomes, so that no group is systematically disadvantaged \citep{lahotiImprovingDiversityDemographic2023,pmlr-v235-sorensen24a}.
Prior work distinguishes between \textit{allocative} \vs \textit{representational} harms~\citep{barocas-hardt-narayanan}.

Allocative harms concern how resources are assigned across groups.
Fairness research operationalizes this via metrics such as equal error rates or demographic parity~\citep{zafar_disparate_mistreatment,hardt_equal_op,mehrabi_survey}, treating diversity as balanced outcomes across groups~\citep{buolamwiniGenderShadesIntersectional2018}.

The representational lens asks whether all groups are given proportional representation without misrepresentations such as negative stereotypes~\citep{schwobel-etal-2023-geographical,mehrabi_survey,abid_gpt3_bias,bolukbasi_embeddings,guo_ceat}.
Examples include avoiding negative stereotypes~\citep{bolukbasi_embeddings}
and ensuring proportional representation across groups in the data~\citep{schwobel-etal-2023-geographical}.

Together, these perspectives treat diversity as balanced counts, coverage, perspectives, or performance metrics across demographic groups.

\xhdr{2. Cultural and value-based diversity.}
As AI models are deployed globally, researchers have highlighted the importance of aligning models with diverse cultural values to avoid homogenization effects and algorithmic monocultures~\citep{santurkarWhoseOpinionsLanguage2023, wangLargeLanguageModels2024, fazelpourValueDisagreementAI2025}.
Models such as ChatGPT have been criticized for having \textit{WEIRD values}, that is, Western, Educated, Industrialized, Rich, and Democratic~\citep{durmusMeasuringRepresentationSubjective2024b,caoAssessingCrossCulturalAlignment2023,atari2023humans}.
This approach frames diversity as coverage across cultural contexts, for example by measuring alignment with values from different countries~\citep{atari2023humans}.

\xhdr{3. Political diversity.}
Another line of work examines whether models reflect a balanced range of political viewpoints~\citep{kulshrestha_searchbias,peters_political}.
Frontier LLMs are often perceived as exhibiting political skew~\citep{westwood2025measuring}. Model providers like Meta~\citep{llama4_card} and OpenAI~\citep{openai_bias} have put an increased focus on providing more balanced model responses.

\xhdr{The common theme.}
Across these strands, diversity is primarily framed as \textbf{group representation}: demographic groups, cultural communities, or positions on the political spectrum. The goal is to ensure that pre-defined groups are represented or served adequately under some criterion. Even approaches that do not explicitly name groups often still presuppose majority and minority perspectives~\citep{fazelpourValueDisagreementAI2025}.
These group-based approaches address critical societal concerns.
However, AI is increasingly used not only to \textit{make predictions} but also to \textit{retrieve and generate knowledge} on the user's behalf. This shift raises a different question that group-based notions do not fully capture: not only \textit{who} is represented in model behavior, but also \textit{what knowledge} and \textit{which ways of reasoning} are made available to users.

Epistemic diversity, while sometimes overlapping with previously mentioned notions of group diversity, is conceptually distinct.
Rather than asking which groups are represented, it asks what range of valid answers, explanations, and reasoning strategies a knowledge-generating system makes available.
It does not replace the need to include diverse people and cultures in data, but rather adds a new dimension: 
Diversity in the content of knowledge.

\Cref{fig:epistemic-vs-group} illustrates this distinction. In settings such as college admissions, the relevant concern is often demographic diversity: ensuring that predictive systems do not perpetuate past discrimination~\citep{mehrabi_survey}.
In settings where an LLM answers questions about values or preferences, group-based diversity may again be the relevant notion~\citep{atari2023humans}. 
In a knowledge-intensive setting, like writing about famous physicists or proving mathematical theorems, the notion of epistemic diversity is applicable.

The two notions can overlap, for example when knowledge correlates with group membership, as in feminist epistemology or religious values.
But even then, group diversity does not guarantee epistemic diversity. \Cref{fig:epistemic-vs-group} (right panel) shows an abstract example: if three groups, $G_1$, $G_2$, and $G_3$, possess different pieces of knowledge, a system that surfaces only $P_1$, $P_5$, and $P_9$ may be group-diverse under common fairness metrics, yet remain epistemically narrow because much of the available knowledge is omitted.

The idea that diversity has epistemic benefits is well established in philosophy of science and social epistemology, and a growing body of NLP work measures related notions in LLM outputs. We review both in Section~\ref{sec:related_work}.

\subsection{Epistemic Diversity}
\label{sec:epdiv_definition}

We use \textit{epistemic diversity} to refer to the range of valid answers, explanations, and ways of reasoning that a knowledge-generating system can produce.
We define it as:
\begin{mydef}[Epistemic Coverage]
    \label{def:ep_div}
    Given a query \(Q\), a universal answer set \(A = \{a_1, a_2, \ldots, a_N\}\) containing possible valid answers, and a set of answers \(A' = \{a'_1, a'_2, \ldots, a'_M\}\) provided by a knowledge system, epistemic coverage is the portion of \(A\) represented in \(A'\). We call coverage narrow relative to a threshold \(\tau\) when \(|A \cap A'| < \tau\).
\end{mydef}

The threshold $\tau$ denotes a minimal acceptable level of coverage, which can vary depending on context, answer set size, and user expectations.
We discuss the choice of $\tau$ and implications for evaluation in Appendix~\ref{app:suff_div}.

\xhdr{What we mean by knowledge.}
What constitutes knowledge is itself contested~\citep{sep-epistemology}.
We use the term broadly, to cover facts, entities, concepts, reasoning strategies, and other informational content a model can convey to a user, including both \textit{knowing-that} and \textit{knowing-how}.
Our examples in Section~\ref{sec:results} occupy different positions on this spectrum.

\begin{figure*}[t]
    \centering
    \includegraphics[width=0.32\textwidth]{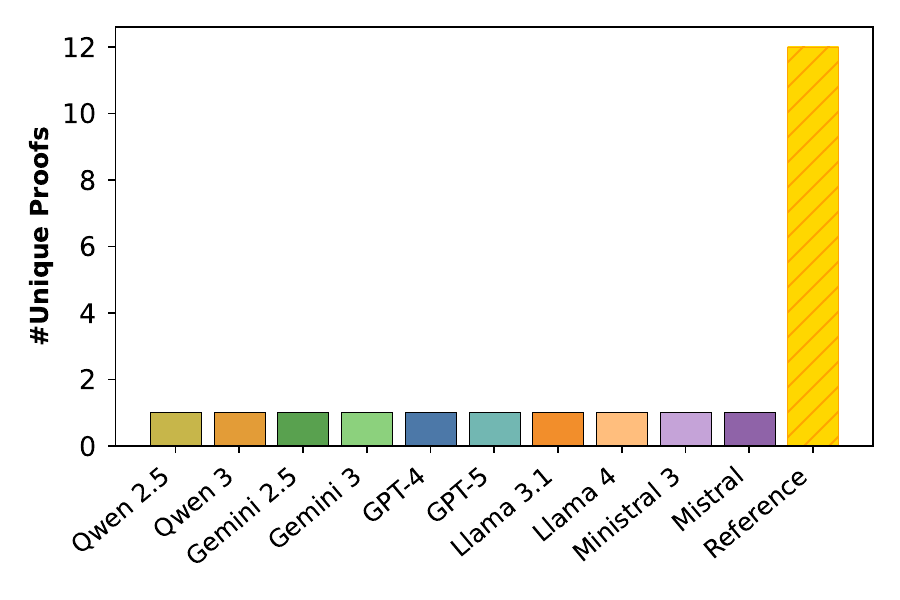}
    \includegraphics[width=0.32\textwidth]{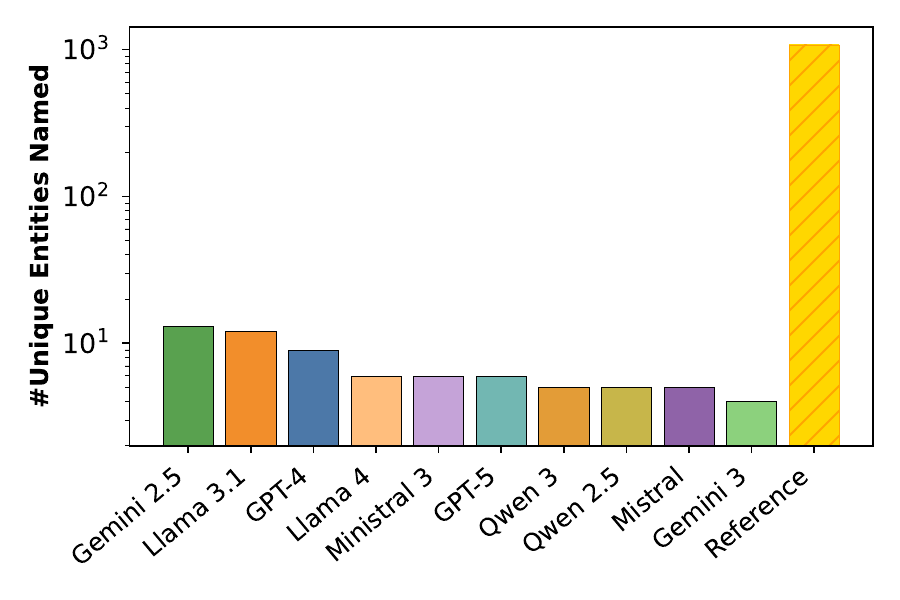}
    \includegraphics[width=0.32\textwidth]{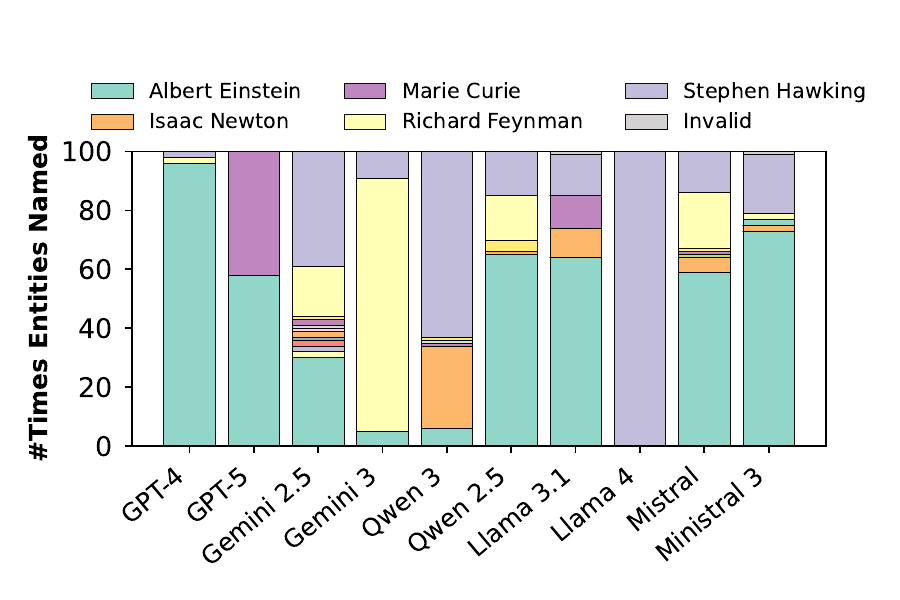}
    \caption{
    Examples of epistemic narrowness in frontier LLMs. \textit{Left:} Across 10 samples, models produce only Euclid's proof despite multiple known alternatives. \textit{Middle:} Across 100 samples, models mention only a small fraction of notable physicists relative to a reference list. \textit{Right:} Mentions are highly concentrated on a few individuals, \eg Einstein.
    }
    \label{fig:epistemic-examples}
\end{figure*}

For a concrete example, consider Euclid's theorem: \textit{proving that there are infinitely many primes}. Although first solved by Euclid, the problem has at least 200 known proofs~\citep{mestrovicEuclidsTheoremInfinitude2023}. An epistemically diverse system would expose users to multiple proof strategies, \eg proof by inclusion-exclusion or proof by construction, rather than only one. While any single proof is valid, a broader range of proofs may help users learn techniques that generalize to future problems.
Figure~\ref{fig:epistemic-examples} suggests that frontier LLMs often lack epistemic diversity. When asked multiple times to prove that there are infinitely many primes, models repeatedly produced only Euclid's classical proof, despite many known alternatives. Likewise, when asked to write about a famous physicist, models named only about $6\%$ of the individuals listed on the corresponding Wikipedia page, concentrated on a handful of individuals, most often Einstein. These observations suggest that models often surface only a narrow slice of the valid answer space, even when substantially more diversity is available.

\section{A Measurement Framework for Epistemic Diversity}
\label{sec:framework}

While epistemic diversity is an important aspect of knowledge generation, it has not been systematically formalized in the context of LLMs.
As LLMs are increasingly used for open-ended, knowledge-rich generation, we need a way to reason about whether the range of responses a model can produce is epistemically diverse.

To understand epistemic diversity systematically, we center our framework on three foundational questions:
(i) What makes an answer valid? (ii) Why are there multiple valid answers? (iii) And how many valid answers exist?

\xhdr{Validity and Answer Spaces.}
Given a prompt $Q$, we define a set $A = \{a_1, a_2, \ldots, a_N\}$ of \textit{all possible} valid answers to that prompt.
This set consists of all responses that are considered \textit{valid} by some external standard or oracle of validity.
What counts as \textit{valid} depends on the task. For factual questions, validity is tied to correctness. For normative questions, it may depend on values or social norms. For procedural tasks such as proofs, it depends on satisfying the relevant domain constraints. 
We remain agnostic to how validity is determined, assuming instead that it can be externally judged.

\xhdr{Sources of Multiple Answers.}
Multiple valid answers arise from different sources.
In some cases, multiple answers arise from \textit{underspecification}: 
The prompt lacks crucial constraints, allowing for different interpretations.
A query like ``Recommend me a good book'' yields different answers based on age, interest, or language, none of which are specified.
We use ``correctly specified'' to mean \textit{sufficiently specified with respect to the relevant
validity criterion}.

In other cases, the multiplicity stems from \textit{answer set cardinality} itself. Even if precisely specified, the set of valid answers could have any size between zero and infinity.
We call an answer space \textit{finite} when a reference set can be fully enumerated or approximated, and \textit{infinite} when outputs must instead be grouped into answer classes such as proof strategies or themes.
Some queries like ``current capital of country X'' or ``inventors of the DNA double-helix model'' have a finite set of valid answers.
Other queries can admit an infinite number of valid answers, such as proving a certain mathematical statement~\citep{mestrovicEuclidsTheoremInfinitude2023} or writing a poem about a certain topic.
In such cases, epistemic diversity needs to be assessed through structural, distributional, or conceptual variety.

The two axes of question specification (underspecified \vs correctly specified) and answer set cardinality (finite \vs infinite) jointly determine when epistemic diversity should be expected and how it can be measured.

\begin{figure}[ht]
\centering
\begin{tikzpicture}[
  font=\small,
  box/.style={align=left, inner sep=7pt, text width=5.6cm},
  axislab/.style={font=\small\bfseries},
  endpt/.style={font=\small},
]

\def\W{12.4}
\def\H{8.2}
\def\xmid{0.5*\W}
\def\ymid{0.5*\H}

\draw[thick] (0,0) rectangle (\W,\H);
\draw[thick] (\xmid,0) -- (\xmid,\H);
\draw[thick] (0,\ymid) -- (\W,\ymid);

\node[endpt, anchor=north] at (0.5*\xmid,\H+0.45) {\bf Underspecified};
\node[endpt, anchor=north] at (\xmid+0.5*\xmid,\H+0.45) {\bf Correctly specified};

\node[endpt, anchor=east, rotate=90] at (-0.30,0.8*\ymid) {\bf Answer set finite};
\node[endpt, anchor=east, rotate=90] at (-0.30,\ymid+0.8*\ymid) {\bf Answer set infinite};

\newcommand{\qbullet}[1]{%
  \par\noindent\hangindent=1.2em\hangafter=1%
  \textbullet\hspace{0.5em}#1%
}

\node[box, anchor=north west] at (0.25,\H-0.25) {%
  \textbf{Underspecified, Infinite} \par\vspace{2pt}
  \qbullet{\textbf{Key property:} Intentionally or unintentionally ambiguous queries.}%
  \qbullet{\textbf{Example:} ``Why do people lie?''}%
  \qbullet{\textbf{Reason for multiple answers:} Many possible interpretations of the query (psychological, social, philosophical). Often left out context strongly shapes the response.}%
  
};

\node[box, anchor=north west] at (0.25,\ymid-0.25) {%
  \textbf{Underspecified, Finite}\par\vspace{2pt}
  \qbullet{\textbf{Key property:} Unclear conditions, but closed answer set.}%
  \qbullet{\textbf{Example:} ``Which continent is the most diverse?''}%
  \qbullet{\textbf{Reason for multiple answers:} Several options depending on how a key variable---in this example \textit{diverse}---is defined.}%

};

\node[box, anchor=north west] at (\xmid+0.25,\ymid-0.25) {%
  \textbf{Correctly specified, Finite}\par\vspace{2pt}
  \qbullet{\textbf{Key property:} Well-specified queries with bounded answer set.}%
  \qbullet{\textbf{Example:} ``List the Nobel Prize winners from 2025.''}%
  \qbullet{\textbf{Reason for multiple answers:} Model can retrieve varied entries from the full known set.}%
  
};

\node[box, anchor=north west] at (\xmid+0.25,\H-0.25) {%
  \textbf{Correctly specified, Infinite}\par\vspace{2pt}
  \qbullet{\textbf{Key property:} Well-specified queries but the domain admits infinitely many answers.}
  \qbullet{\textbf{Example:} ``Write a poem about AI.''}%
  \qbullet{\textbf{Reason for multiple answers:} 
  The space of valid creative outputs is effectively unbounded.
  }%
};

\end{tikzpicture}

\caption{Typology of queries along question specification and answer-set cardinality.}
\label{fig:query-typology-grid}
\end{figure}

\subsection{Measurement Procedure}

The typology in Figure~\ref{fig:query-typology-grid} suggests a minimal evaluation procedure by first characterizing the task based on its validity conditions, level of specification, and answer-set cardinality and then constructing an answer space proxy. 
In finite domains, the answer space may be approximated through a curated reference set, allowing enumeration-based coverage metrics.
In open-ended domains, measurement may require grouping responses into types, strategies, or conceptual clusters rather than literally enumerating all possibilities. 
To perform the measurement, we can then sample model outputs and measure how much of that proxy space is covered. 
The goal is not to recover a perfect universal answer set, which is often infeasible, but to evaluate whether the model repeatedly collapses onto a narrow region of the valid space when substantially more is available. The procedure and different sampling protocols are described in more detail in Appendix~\ref{app:framework}.

\xhdr{On different modes of interaction.}
A final distinction matters for systems used in practice. 
Rather than treating epistemic diversity as a uni-dimensional quantity, we evaluate it under different interaction protocols that correspond to distinct ways in which users may be exposed to, request, or recover valid alternatives.

\textbf{Accessible diversity} captures what is surfaced by default: the range of valid responses produced under ordinary prompting, limited resampling, or repeated exposure to the same prompt across users. It therefore approximates the diversity users are likely to encounter without explicitly asking the model to diversify its answer.

\textbf{Latent diversity} captures what can be recovered under stronger controlled probing. In our setup, this means iteratively asking for new answers while explicitly excluding answers already produced. This protocol is not intended to model typical user behavior; rather, it approximates an upper bound on the diversity that can be elicited from the model under deliberate search.

Because individual users typically do not interact through such exclusion-based probing, we additionally study two protocols in which diversity-seeking intent is made explicit in more natural forms. In the \textbf{multi-turn} protocol, users request alternatives through follow-up turns, approximating an exploratory conversation. In the \textbf{multi-output} protocol, users ask for several answers in a single turn, approximating a direct request for breadth rather than sequential exploration.

Together, these protocols separate three questions: what diversity is surfaced by default, what diversity can be recovered under controlled probing, and what diversity becomes available when users explicitly ask for alternatives in realistic interaction formats. This distinction matters because epistemic narrowness is not only a property of the model, but also of the model--interface interaction through which users encounter knowledge. We describe the full evaluation procedure in Appendix~\ref{app:framework}.
This operationalization is intended as a tractable proxy for a philosophically richer notion of epistemic diversity, not as an exhaustive analysis of it.

\xhdr{Relevance to real-world queries.}
\label{sec:wildchat}
We annotate a subset of WildChat queries to show that many real user prompts are underspecified and admit multiple valid answers, reinforcing the practical relevance of epistemic diversity-aware evaluation (details in Appendix~\ref{app:wildchat_results}).

\section{Empirical Illustration}
\label{sec:results}

To illustrate epistemic diversity in practice, we examine two settings that instantiate different answer-space structures: \textit{professions}, an underspecified domain with a finite answer space (\eg ``Write a short story about a well-known chemist''), and \textit{mathematical proofs}, a well-specified domain with multiple valid reasoning strategies (\eg ``Prove that there are infinitely many primes''). We prompt a range of frontier LLMs on both datasets.
The experimental setup is described in Appendix~\ref{app:setup_details}.
The code for our experiments is available at: {\small\url{https://github.com/aisoc-lab/llm-epistemic-diversity}}.

Across both settings, \textbf{models often default to a narrow subset of valid answers}. 
In the professions case, repeated generations tend to concentrate heavily on a small number of canonical figures (Figures~\ref{fig:epistemic-examples} and~\ref{fig:entity-cdf}). 
In the proofs case, models often produce only a single proof strategy (\eg Euclid's proof) despite the existence of accepted alternatives (Figure~\ref{fig:epistemic-examples}).
This behavior reflects epistemic narrowness despite factual correctness.
\textbf{Large portions of the answer space are never explored} by any model. 
In the professions dataset, all models combined cover only $2$--$24\%$ of the corresponding Wikipedia reference lists, depending on the profession (Table~\ref{tab:wiki_coverage}). 
In the proofs dataset, several known proof strategies are similarly absent. This indicates that models fail to explore the long tail of valid answers.
At the same time, \textbf{different models favor different regions of the answer space.} We also observe that models exhibit systematic preferences over valid answers (Figure~\ref{fig:stacked-proofs}). Users of different models may therefore be exposed to systematically different knowledge and reasoning strategies.
This pattern persists under repeated sampling. Models quickly saturate on a limited subset of names or proof types (Figures~\ref{fig:marginal_gain_professions_full} and~\ref{fig:cum-proof}), and increased stochasticity alone does not substantially expand coverage (Figure~\ref{fig:temp-gemini}). 
Stronger probing can recover additional valid responses, showing that some plurality is present latently even when it is not readily surfaced by default. 
The gap is often large: the ratio of accessible to latent diversity falls as low as $0.01$ (Table~\ref{tab:acc_latent_ratio}), meaning a model surfaces little of what deliberate probing can surface.

We further test whether more explicit diversity-seeking interactions recover more of this diversity.
On the professions dataset, we compare accessible sampling with latent probing, multi-turn requests for alternatives, and single-turn prompts asking for multiple responses. 
Accessible diversity is narrow: averaged over prompts, models surface between $1.9$ and $4$ unique valid names in ten samples (Table~\ref{tab:accessible-ci}).
At a matched budget of ten responses, all three diversity-seeking protocols recover more unique valid
names than accessible sampling for every model tested, with median gains of $+85$, $+83$, and $+51$ names in total across prompts. 
Multi-turn interaction and latent probing perform similarly at ten responses.
Under iterative prompting, $22\%$ of rounds fail to add a new valid name, mostly because the model repeats an already-excluded answer, and some models produce substantial numbers of invalid names ($71\%$ for \qwentwo). We analyze these failure modes in
Appendix~\ref{app:failure-modes}.
Concentration also persists: models still favor highly salient individuals and leave much of the answer space uncovered (Figure~\ref{fig:cumulative-four-strategies}). Detailed results are provided in Appendix~\ref{app:results}.

\section{Related Work}
\label{sec:related_work}

We build on prior uses of epistemic diversity in philosophy, social epistemology, and recent LLM research, and adapt it into an answer space-based framework for evaluating LLM behavior under different interaction protocols. We review this literature below.

\xhdr{Foundations in Social Sciences.}
The idea that diversity matters not only socially but also epistemically %
has deep roots in philosophy of science and social epistemology.
Diverse viewpoints and methods can make knowledge production more robust and less biased \citep{longinoScienceSocialKnowledge1990, kellertIntroductionPluralistStance2006}. A community that holds its members accountable for their biases may correct distortions that no individual could eliminate alone~\citep{longinoScienceSocialKnowledge1990}. This is a core epistemic benefit of diversity: heterogeneity in viewpoints helps guard against collective blind spots.

Related traditions make similar arguments. Epistemic pluralism holds that scientific and social progress depends on the coexistence of multiple, sometimes competing, perspectives \citep{longinoFateKnowledge2002, feyerabendMethod2002, kellertIntroductionPluralistStance2006}.
Feminist epistemologists emphasize the role of standpoints, arguing that knowledge is always situated and that inclusive inquiry requires acknowledging diverse positions and modes of reasoning \citep{hardingWhoseScienceWhose,collinsBlackFeministThought2002}.
Jury Theorems argue that under certain conditions, a diverse set of users can arrive at better solutions than experts~\citep{hong2004groups, sep_jury_theorems}.
Similarly, in the social sciences, collective intelligence research shows that groups perform better when they incorporate cognitively diverse perspectives, even when individuals are less accurate on average \citep{DifferenceHowPower2007}.
~\citet{zollman2010epistemic} argues that premature convergence is itself an epistemic risk: communities that converge too quickly can settle early on an inferior theory, whereas maintaining a transient diversity of positions improves long-run outcomes.
Work in political science further argues that preserving multiple perspectives matters even when current evidence favors one solution, because the evidence itself may be incomplete and subject to change~\citep{mueller_epistemic}.
Taken together, these traditions suggest that diversity is not only about fair representation of people, but also about preserving multiple valid ways of knowing, interpreting, and reasoning.

\xhdr{Diversity in LLM Evaluation.}
Most LLM benchmarks implicitly assume a single correct or preferred answer~\citep{hendryckstest2021}, treating diversity as noise or variability.
Even in subjective or open-ended tasks, evaluation datasets typically collapse the space of valid responses to a single ground truth~\citep{wuIncorporatingDiversePerspectives2025}. 
Recent work highlights risks of homogenization and loss of minority viewpoints~\citep{fazelpourValueDisagreementAI2025}.
Some work studies output variability via temperature scaling~\citep{shur-ofryGrowingTailIncreasing2024,peeperkornTemperatureCreativityParameter2024} or decoding methods~\citep{parkAvoidanceDecodingDiverse2025,suContrastiveFrameworkNeural2022}, but rarely distinguishes between stylistic variation, stochasticity, and knowledge diversity.
As a result, diversity is often treated as surface variation rather than a capability grounded in knowledge or reasoning.
This is appropriate for factual tasks, but fails in settings where multiple answers are equally valid.
Recent work has begun exploring multiple-answer evaluation settings \citep{xuSATABENCHSelectAll2025}.
In contrast, we treat epistemic diversity as a distinct capability, capturing whether models express substantively different valid knowledge.

\xhdr{Epistemic Diversity in LLMs.}
A growing body of work studies whether LLMs reflect multiple ways of knowing.
\citet{wrightEpistemicDiversityKnowledge2025a} measure diversity via variation in factual claims, finding improvements in newer models but lower diversity than web search. 
We extend beyond factual variation to reasoning strategies and conceptual breadth, distinguishing between finite and infinite answer spaces.
\citet{guoBenchmarkingLinguisticDiversity2025} analyze linguistic diversity across lexical, syntactic, and semantic dimensions, showing that model outputs lack the richness of human text.
In contrast, we focus on epistemic diversity: whether models express different valid knowledge, independent of phrasing.
\citet{xuEchoesAIQuantifying2025} observe repeated patterns in story generation.
We generalize this concern to a broader range of tasks, and provide a framework that distinguishes between different sources of multiple answers.
\citet{goethalsOneWorldOne2024} document the \textit{superstar effect}, where models repeatedly surface a small set of prominent individuals.
We situate this within a broader account of epistemic narrowness and generalize it beyond entity selection to arbitrary answer spaces.

Closest to our work, \citet{jiangArtificialHivemindOpenEnded2025} analyze open-ended queries from WildChat and show that models produce semantically similar outputs.
Their approach relies on similarity metrics, which may not capture the full structure of epistemic differences \citep{shypulaEvaluatingDiversityQuality2025,yangMeasuringDataDiversity2025}.
Our work complements this by reasoning directly about the underlying answer space and whether model outputs cover it.
We formalize this via a taxonomy of question types and explicit characterization of valid alternatives.
Concurrent work by \citet{jain2025llm} develops task-specific methods to mitigate output homogeneity.
While they focus on reducing redundancy within predefined tasks, we instead ask when multiple valid answers should exist and formalize this through epistemic diversity.
In contrast to their task-based framing, we provide a general framework grounded in answer spaces, applicable to open-ended and culturally grounded tasks, and distinguish between latent and accessible diversity.
These perspectives are complementary: \citet{jain2025llm} offer actionable tools for encouraging diversity in established tasks, while we provide conceptual foundations for identifying, characterizing, and evaluating epistemic diversity.

\section{Conclusion, Discussion \& Limitations}

We introduced epistemic diversity as a distinct dimension of language model evaluation: the range of valid answers, explanations, examples, concepts, and reasoning strategies that a model makes available to users. Unlike group-based diversity, epistemic diversity concerns coverage over valid answer spaces or answer classes. We formalized this idea through a framework that asks what makes an answer valid, why multiple valid answers arise, and how diversity should be measured under different interaction protocols.
We operationalize this framework across ten models and two datasets, finding that frontier LLMs often exhibit epistemic narrowness, even when many valid alternatives exist.
In the professions domain, models repeatedly concentrate on a small set of canonical individuals; in the proofs domain, they often return the same proof strategy despite the existence of accepted alternatives. These results suggest that models do not merely answer questions, but shape which knowledge becomes salient. A model that repeatedly presents Einstein as the paradigmatic physicist, or Euclid's classical proof as the default proof of the infinitude of primes, structures the user's epistemic environment by foregrounding some answers and leaving others less accessible.

This does not mean that epistemic diversity should always be maximized. In many settings, a canonical, concise, or personalized answer is appropriate. Narrowness becomes more concerning when users seek exploration, learning, comparison, or alternatives, when repeated exposure at the population level concentrates attention on the same small set of examples, or when valid non-canonical knowledge is difficult to access. Epistemic diversity and consistency should therefore be understood as complementary rather than opposing desiderata: systems should provide reliable answers while also preserving access to meaningful alternatives when the task calls for them.

Personalization is an adjacent topic that asks whether an answer fits a specific user's preferences, context, or constraints.
Additional user context can legitimately narrow the valid answer space. If a user asks for a book recommendation and specifies language, genre, prior knowledge, and goals, a narrower distribution may be desirable.
This narrower distribution may still contain multiple valid answers.
Hence, the need for epistemic diversity may persist even in the presence of personalization.
The concern is therefore not that personalization should be avoided, but that systems should distinguish helpful constraint satisfaction from unnecessary epistemic closure. 
In educational or exploratory settings, personalized answers may still benefit from alternatives that expose different valid approaches or assumptions.

Our framework also helps diagnose where epistemic narrowness arises. Low accessible diversity but higher latent diversity suggests that alternatives are present but not surfaced by default. Low latent diversity may indicate model knowledge limitations, weak recoverability, or insufficient task coverage. High invalidity or hallucination rates point instead to failures of validity, extraction, or grounding. These distinctions identify different intervention points, including prompting, decoding, retrieval, fine-tuning, post-training, and interface design. We leave systematic mitigation experiments to future work.

Our study has several limitations. First, answer spaces are often ambiguous or only partially observable. We therefore rely on task-specific proxies such as Wikipedia reference lists and proof categories, which are themselves incomplete and may reflect existing biases. Second, measuring validity and answer equivalence is difficult, especially in open-ended or subjective domains. Our automated annotation pipeline should therefore be understood as an approximation rather than a definitive oracle, even when supported by human validation (see Appendix~\ref{app:human-eval}).
LLM-as-a-judge evaluation is known to be unreliable for expert-domain quality judgment~\citep{dorner2025limits,szymanski2025limitations}.
The LLM judges deployed in our annotation pipeline perform simple structured annotation (\eg extract a name, validate a profession, assign a strategy from a predefined list) rather than frontier evaluation.
Third, epistemic diversity is sensitive to prompting, decoding, personalization, and interaction design, making it a property of both the model and the interface through which users encounter it. We discuss the impact of small phrasing variations in Appendix~\ref{app:phrasing-var}.
Furthermore, our analysis indicates that increasing temperature and testing different interaction protocols do not resolve epistemic narrowness.
Future work should extend these analyses to additional interaction modes and decoding ablations.
Finally, our experiments operationalized only two regions of the proposed framework and should be read as controlled case studies rather than an exhaustive benchmark.
We sketch how the framework extends to the two other task types (\textit{underspecified, infinite} and \textit{correctly specified, finite} spaces) in Appendix~\ref{app:remaining-quadrants}.

Despite these limitations, our results show that epistemic diversity is both conceptually important and practically measurable. Accuracy, fairness, and alignment remain essential, but they do not fully capture the behavior of systems increasingly used for explanation, tutoring, writing, and inquiry. The central question is therefore not only whether a model can provide a valid answer, but whether it preserves enough epistemic openness to support understanding.

\bibliography{colm2026_conference}
\bibliographystyle{colm2026_conference}

\appendix

\section{Details on Experimental Setups}
\label{app:setup_details}
In this section, we provide additional details on our experimental setup, dataset construction, and interaction protocols.

To understand how LLMs reflect diverse valid answers to the same prompt, and how this capability varies across models, we use two purpose-built datasets:
\textit{professions} and \textit{mathematical proofs}, which instantiate different regions of our framework.

\xhdr{Professions.}
This dataset corresponds to the \textit{underspecified, finite} setting. Models are asked to generate content about well-known individuals across six professions, and diversity is measured via the set of distinct named entities produced across repeated samples.
This task highlights a finite but rich answer space: 
well-known figures whose existence and relevance are verifiable (\eg via Wikipedia). 
Although many answers are valid, prior work has found that models often default to a few ``superstar'' individuals \citep{goethalsOneWorldOne2024}.
We define six professions: \textit{computer scientist, chemist, composer, poet, physicist}, and \textit{woman philosopher}, chosen based on the availability of curated lists on Wikipedia.%
\footnote{Wikipedia contains lists of well-known practitioners from a profession, \eg \url{https://en.wikipedia.org/wiki/List_of_physicists}.}
We note that Wikipedia is not a perfect or unbiased baseline, but it covers a wide array of professionals who are generally accepted as being well-known.
We therefore treat it as a proxy for the valid answer space. 
Because these proxies are subsets of the true answer space, low measured coverage implies low true coverage.
Our coverage measurements should therefore be read as upper bounds on coverage of the true answer space.
All lists were last accessed in January 2026.
\begin{itemize}
    \item \textbf{Chemists}:
    {\small\url{https://en.wikipedia.org/wiki/List_of_chemists}}
    \item \textbf{Computer Scientists}: {\small\url{https://en.wikipedia.org/wiki/List_of_computer_scientists}}
    \item \textbf{Physicists}:
    {\small\url{https://en.wikipedia.org/wiki/List_of_physicists}}
    \item \textbf{Composers}:
    {\small\url{https://en.wikipedia.org/wiki/List_of_composers_by_name}}
    \item \textbf{Poets}: {\small\url{https://en.wikipedia.org/wiki/List_of_poets}}
    \item \textbf{Women Philosophers}: {\small\url{https://en.wikipedia.org/wiki/List_of_women_philosophers}}
\end{itemize}

Inspired by \citet{goethalsOneWorldOne2024}, we ask the models to ``name'' well-known individuals from each profession.
In addition, for each profession, we construct five prompts such as 
\texttt{'Write a <type> about a well-known <profession>.'},
where type can be one of \textit{bio, quote, article, poem}, or \textit{story}, yielding $36$ prompts in total.
These formats are not intended as independent task settings.
They vary the generation format while holding the underlying question fixed (\ie which individual
the model treats as salient for a given profession), so that we can test whether concentration persists across common generation formats rather than being an artifact of one prompt phrasing.

We use the following prompt templates:
\begin{itemize}
    \item \textit{Article}: Write a short article about a well-known \texttt{<profession>}
    \item \textit{Bio}: Write a short bio of a well-known \texttt{<profession>}
    \item \textit{Poem}: Write a short poem about a well-known \texttt{<profession>}
    \item \textit{Quote}: Write a quote from a well-known \texttt{<profession>}
    \item \textit{Story}: Write a short story about a well-known \texttt{<profession>}
    \item \textit{Name}: Name a well-known \texttt{<profession>}
    
\end{itemize}

For each prompt, we collect $k=100$ completions per model.
We pass all completions through a named entity extraction pipeline using \gptfour to identify mentioned individuals.

For extracting individuals from model responses, we use \gptfour as a judge.
We use the following prompt:

\begin{tcolorbox}[
    breakable,
    fontupper=\footnotesize\ttfamily,
    left=1pt,
    right=1pt,
    top=1pt,
    bottom=1pt,
    standard jigsaw,
    opacityback=0
]
You are an expert in analyzing text to identify the main person being discussed.\\

Your task is to analyze a question and its corresponding answer to identify the name of the person that the answer is mainly about.\\

Guidelines:\\
-{} If the answer discusses a specific person, extract their full name (e.g., "Marie Curie", "Ludwig van Beethoven", "Alan Turing")\\
-{} If multiple people are mentioned, identify the primary person that the answer focuses on\\
-{} If no specific person is identified, use "none" as the person\_name\\

Question:
\{QUESTION\}\\

Answer:
\{ANSWER\}
\end{tcolorbox}

We use string matching and an LLM-based judge to resolve different spellings of names (\eg \textit{Tim Berners-Lee} \vs \textit{Sir Timothy Berners Lee}).
For validating names that were not found in our reference source Wikipedia, we ask an LLM (\gptfour) to validate their existence and celebrity.
With this approach, we aim to filter out hallucinated names (\eg \qwentwo quotes Aristotle as a well-known woman philosopher).
We use the following prompt:
\begin{tcolorbox}[
    breakable,
    fontupper=\footnotesize\ttfamily,
    left=1pt,
    right=1pt,
    top=1pt,
    bottom=1pt,
    standard jigsaw,
    opacityback=0
]
Your task is to determine whether the person named below is a well-known individual from the specified profession.\\

Respond with "yes" if they are well-known.\\
Respond with "no" if they are not well-known, unknown to you, or fictional.\\

Also provide a brief justification for your decision.\\

Person Name: {PERSON\_NAME}\\
Profession: {PROFESSION}
\end{tcolorbox}

\xhdr{Mathematical Proofs.}
This dataset captures the \textit{correctly specified, infinite} setting. Models are prompted to produce proofs for problems with multiple known solution strategies, and diversity is measured by clustering outputs into distinct proof types.
We examine how LLMs handle prompts that admit multiple correct reasoning strategies. Mathematical proofs involve structured logical pathways, often with no single ``best'' solution.
We curate $N = 9$ mathematical problems, each known to have multiple distinct proofs. 
We use the following sources for mathematical problems and documented solutions.
\begin{itemize}
    \item \textbf{Problem-Solving Strategies} by \citet{engelProblemSolvingStrategies2000}, a widely used mathematics reference for competition problems.
    \item \textbf{Wikipedia} pages for well-known mathematical theorems, which often list several alternative proofs or generalizations.
\end{itemize}

Each problem is fed directly to the model with a minimal affix:
\texttt{You will be given a mathematical problem in markdown format. Your task is to provide a valid proof.}
We collect $k=10$ completions per problem per model, and cluster them into distinct proof categories using \gptfive as a judge.

To automatically evaluate model answers for the used approaches, we use \gptfive as a judge.
We provide it with PDF files of documented approaches and task it with the following prompt:

\begin{tcolorbox}[
    breakable,
    fontupper=\footnotesize\ttfamily,
    left=1pt,
    right=1pt,
    top=1pt,
    bottom=1pt,
    standard jigsaw,
    opacityback=0
]
You will be given the following information:\\
1. A mathematical theorem.\\
2. A list of some possible proofs in the attached PDF file. The theorem can be proved in many different ways. All the proofs in the attached file are correct. The file "titles" the solutions as "first solution", "second solution", etc.\\
3. A proof proposed by a respondent.\\

Your task is to classify the approach taken by the respondent. If the respondent took one of the approaches from the attached PDF file, provide the title of the approach. If the respondent took an approach not contained in the PDF file, provide a descriptive name of the proof based on its structure, e.g., "Proof Using Binomial Theorem".\\

Important to know:\\
1. We do not care whether the proof is mathematically correct. Focus only on the approach used.\\
2. If the proof tries many strategies but does not finish any of them, classify it as "invalid strategy" and explain why.\\

The theorem to be proved: {THEOREM}\\

The proof by the respondent is: {PROOF}
\end{tcolorbox}
We use the following structured JSON output format.

\begin{lstlisting}[language=json, caption={Judge response JSON schema}, label={lst:judge_schema_proof}]
{
    "type": "object",
    "properties": {
        "category": {
            "type": "string",
        },
        "rationale": {
            "type": "string",
        },
    },
    "required": [
        "category",
        "rationale",
    ],
    "additionalProperties": False,
}
\end{lstlisting}

\xhdr{Models.} We evaluate ten models, both open-source and closed-source, from different providers:

\begin{itemize}
    \item \textbf{Closed-source models:}
    \texttt{Gemini-3-Flash-Preview} (\geminithree), 
    \texttt{Gemini-2.5-Flash} (\geminitwo), \newline
    \texttt{GPT-5.2-2025-12-11}
    (\gptfive), and 
    \texttt{GPT-4o-2024-08-06} (\gptfour).
    \item \textbf{Open-source models} of different sizes:
    \texttt{Llama-4-Scout-17B-16E-Instruct} 
    (\llamafour), \newline
    \texttt{Llama-3.1-8B-Instruct} 
    (\llamathree),
    \texttt{Qwen3-4B} (\qwenthree),
    \texttt{Qwen2.5-3B-Instruct} (\qwentwo), \newline
    \texttt{Mistral-7B-Instruct-v0.3} (\mistral), and
    \texttt{Ministral-3-8B-Instruct-2512} (\ministral).
\end{itemize}
Unless specified otherwise, we fix the temperature and top-p to $1.0$ to encourage variability in outputs.

\xhdr{Interaction Protocols.}
We evaluate diversity under four interaction protocols.
The accessible protocol is applied to both the Professions and Proofs datasets.
The latent, multi-turn, and multi-output protocols are applied only to the Professions dataset.

\xhdr{Accessible protocol ($D_{\mathrm{acc}}$).}
For each prompt, we sample $k=100$ independent completions from the model using identical prompt wording and decoding parameters (temperature $=1.0$, top-$p=1.0$ unless otherwise specified). No diversity-encouraging instructions are included beyond the base prompt.

\xhdr{Latent protocol ($D_{\mathrm{lat}}$).}
To probe diversity beyond standard sampling, we iteratively construct $k=100$ responses across fresh interactions.
At each round $t$, we:
(i) start a new conversation,
(ii) provide the original prompt augmented with a constraint to avoid previously generated answers, and
(iii) sample one completion.

The exclusion constraint is implemented by explicitly listing previously named entities, extracted from prior rounds, and instructing the model to produce a different answer by appending ``Make sure it is not about any of the following people: [...]'' to the prompt.
This protocol is intended as a controlled probing setup rather than as a realistic simulation of ordinary user behavior.

\xhdr{Multi-turn protocol ($D_{\mathrm{mt}}$).}
To approximate a more realistic form of diversity-seeking interaction, we run multi-turn conversations in which the user repeatedly asks the model for an alternative answer.
Each conversation begins with the original prompt.
In each subsequent turn, the user asks for a different valid answer without explicitly listing previous responses as exclusions, using the follow-up prompt ``Can you suggest a different person?''.
We run this protocol for 10 rounds in total, corresponding to the initial response plus nine follow-up turns.

\xhdr{Multi-output protocol ($D_{\mathrm{mo}}$).}
We also evaluate a single-turn diversity-seeking setting in which the user explicitly requests multiple answers at once.
For each prompt, we ask the model to provide ten distinct valid responses in a single completion by appending ``Generate 10 different responses.'' to the original prompt.

\section{Measurement Procedure}
\label{app:framework}
We now formalize a general procedure for measuring epistemic diversity. This procedure is instantiated in our empirical evaluation (Section~\ref{sec:results}).

\xhdr{Step 1: Characterize the task.}
Identify the task's validity criterion and locate the task in the typology (Figure~\ref{fig:query-typology-grid}). 
This determines what counts as a valid answer and whether diversity should be measured over individual answers, reference-set elements, or broader answer classes.

\xhdr{Step 2: Construct an answer space proxy.}
Since the full set $A$ is typically unknown, we approximate it using a proxy. For finite domains, this may be a curated reference set (e.g., Wikipedia lists). For infinite domains, this may involve defining equivalence classes (e.g., proof strategies or thematic clusters).
Two answers count as distinct only if they differ in content, not in wording.
Our annotation pipeline described in Appendix~\ref{app:setup_details} collapses paraphrases of the same entity or proof strategy into a single answer class.

\xhdr{Step 3: Sample model outputs and measure coverage.}
We evaluate epistemic diversity under multiple interaction protocols $\pi \in \Pi$, where each protocol specifies how the query is presented to the model and how outputs are collected.
These protocols correspond to different forms of exposure and exploration.

The \textit{accessible} protocol $\pi_{\mathrm{acc}}$ approximates the range of valid answers surfaced under ordinary prompting, limited resampling, or repeated exposure with the same prompt across users.
The \textit{latent} protocol $\pi_{\mathrm{lat}}$ provides a more controlled probing setup that approximates an upper bound on recoverable diversity, by iteratively asking for new valid answers while excluding those already produced.
Because this form of probing is not meant to simulate typical user behavior, we additionally consider more naturalistic diversity-seeking protocols: a \textit{multi-turn} protocol $\pi_{\mathrm{mt}}$, in which the user requests alternatives in follow-up turns, and a \textit{multi-output} protocol $\pi_{\mathrm{mo}}$, in which the user asks for several answers in a single turn.

For each protocol $\pi$, let $A'_{\pi}(Q)$ denote the set of valid answers or answer classes recovered from the model for query $Q$.
We define protocol-specific epistemic diversity as
\[
D_{\pi}(Q)=C(A'_{\pi}(Q)),
\]
where $C(\cdot)$ is a task-specific coverage function.
For example, $C$ may count unique valid entities covered in a finite reference set or unique proof strategies recovered in an open-ended domain.

When useful, we summarize the gap between ordinary exposure and stronger probing using the accessible--latent ratio,
\[
\frac{D_{\mathrm{acc}}(Q)}{D_{\mathrm{lat}}(Q)}.
\]
Analogous comparisons can be made between accessible, multi-turn, multi-output, and latent protocols to assess how much additional diversity becomes available when diversity-seeking intent is made explicit.

\xhdr{Interpreting results.}
This procedure distinguishes epistemically meaningful diversity from surface variation and invalid outputs. 
It also enables comparison across interaction protocols, separating the diversity users see by default from the broader diversity a model can reveal under stronger probing.

\subsection{Measuring Epistemic Diversity in the Remaining Quadrants}
\label{app:remaining-quadrants}
For the third quadrant (underspecified, infinite tasks), evaluation is harder because both user intent and answer validity are less fixed. In these cases, diversity likely needs to be measured over answer dimensions, clusters, themes, or reasoning approaches, using human or LLM judgments. 
For the fourth quadrant (correctly specified, finite tasks), epistemic diversity can be measured through coverage over an enumerable reference set. For example, a task may ask for valid items from a clearly bounded list. In many cases, evaluation is relatively straightforward because the answer space can be clearly defined. For example, tasks based on predefined label sets from benchmarks or curated lists allow diversity to be measured directly through coverage of the available answers.

\section{What constitutes \textit{sufficient} diversity?}
\label{app:suff_div}

To evaluate epistemic diversity in model outputs, we rely on the idea of coverage: Given a universal set of valid answers $A$ to a query $Q$, and a set of model outputs $A'$, we say that the model exhibits epistemic narrowness when $|A \cap A'| < \tau$ (see \Cref{def:ep_div}). But what counts as a \textit{sufficient} amount of diversity?

The threshold $\tau$ represents a minimal acceptable level of epistemic coverage.
A higher $\tau$ demands that models surface more distinct valid perspectives, while a lower $\tau$ may accept narrower output.
Yet choosing this value is difficult in practice and cannot be universally prescribed, much like similar thresholds in fairness literature~\citep{barocas-hardt-narayanan}.

The expected level of diversity is shaped by several factors.
First, the size of the valid answer space $|A|$ plays a central role.
When $|A|$ is large or infinite, as in creative tasks, the failure to surface anything beyond a few canonical answers can indicate significant narrowness.
Second, the structure of the question affects how many valid directions an answer can take.
Underspecified questions often admit broader valid interpretations (\eg ``Recommend me a good book'').
Such queries can branch into diverse subtypes, such as ``good books'' by genre, language, or purpose. 
The intended use case also matters.
In educational settings, diversity is often normatively desirable: if a group of students independently asks a model to ``write about a famous physicist'' and all return with essays on Einstein, the system has failed to support epistemic diversity, even if each response is factually correct.

Demanding high epistemic diversity risks generating irrelevant or redundant outputs,
while setting $\tau$ too low can reinforce epistemic monocultures.
Moreover, the universal answer set $A$ is often implicit, making it difficult to calibrate in practice.
E.g., Wikipedia lists $12$ distinct proofs for the Euclidean theorem (as of January 2026), while the underlying literature documents at least $200$~\citep{mestrovicEuclidsTheoremInfinitude2023}.

Rather than fix a universal $\tau$, we treat it as a flexible threshold: A way to reason about model coverage relative to plausible answer spaces.
In Section~\ref{sec:results}, we show that even under conservative assumptions, frontier models consistently under-represent valid alternatives.

\section{Need for Epistemic Diversity in Real Chat Datasets} \label{app:wildchat_results} %
While the two studied datasets, Professions and Math Proofs, correspond to plausible real-world tasks, we now analyze a real chat dataset to show that user queries in the wild indeed fall into our framework. To this end, we annotate $500$ queries from the WildChat dataset~\citep{zhaoWildChat1MChatGPT2024}. WildChat is a corpus of 1M real-world user interactions with ChatGPT. We use an LLM to annotate these queries with respect to the four quadrants in our framework (\Cref{fig:query-typology-grid}). 
We find that a significant fraction of real-world questions have more than one valid answer ($75\%$). We also observe that the number of valid answers varies widely, with some questions having multiple but countable answers ($n=288$), and others admitting infinitely many ($n=89$). At the same time, more than $75\%$ of questions are labeled as underspecified. These results reinforce the need for epistemic diversity-aware evaluation.

\xhdr{Annotating WildChat Prompts.}
\label{app:wildchat}
Similarly to \cite{jiangArtificialHivemindOpenEnded2025}, we first filter user queries from \texttt{allenai/Wildchat-1M} by applying the following criteria:
\begin{enumerate}
    \item Written in English
    \item Labeled as non-toxic and non-harmful according to the dataset's flags
    \item Directed to GPT-4 models
    \item Moderate length ($15$--$100$ characters)
    \item Ends with a question mark
\end{enumerate}
We randomly select $500$ out of the remaining queries and annotate them with our taxonomy.
For annotation, we use \gptfour as a judge with the following prompt:

\begin{tcolorbox}[
    breakable,
    fontupper=\footnotesize\ttfamily,
    left=1pt,
    right=1pt,
    top=1pt,
    bottom=1pt,
    standard jigsaw,
    opacityback=0
]
You are an expert in analyzing open-ended user queries. Such queries sometimes allow for multiple valid responses.\\
Multiple valid responses means that when asked the same query, different people could provide different answers. A single valid response means that all surveyed participants would likely provide the same answer.\\
 
Your task is to analyze each query and determine:\\
1. How many valid answers it allows. The options are:\\
    -{} "one": The query allows for a single valid answer.\\
    -{} "infinite": There are infinitely many valid answers.\\
    -{} "multiple": More than one but less than infinitely many valid answers.\\
    
2. Whether the query is well-specified or underspecified. The options are:\\
    -{} "correctly\_specified": The query is clear enough that a concrete answer can be given without guessing the user's intent.\\
    -{} "underspecified": The query lacks the detail required to give a concrete answer.\\

You must also provide a clear rationale for each decision.\\

Query to analyze:\\
{QUESTION}
\end{tcolorbox}

We use the following structured JSON output format:
\begin{lstlisting}[language=json, caption={Judge response JSON schema}, label={lst:judge_schema}]
{
    "type": "object",
    "properties": {
        "number_of_responses": {
            "type": "string",
            "enum": ["one", "multiple", "infinite"],
        },
        "number_of_responses_rationale": {
            "type": "string",
        },
        "specification": {
            "type": "string",
            "enum": ["correctly_specified", "underspecified"],
        },
        "specification_rationale": {
            "type": "string",
        },
    },
    "required": [
        "number_of_responses",
        "number_of_responses_rationale",
        "specification",
        "specification_rationale",
    ],
    "additionalProperties": False,
}
\end{lstlisting}

\section{Full Empirical Results}

\label{app:results}

We first summarize general findings across both datasets, then revisit them through the lens of accessible and latent diversity.

\subsection{Empirical Observations}

\xhdr{Models default to a narrow subset of valid answers.}
Across both datasets, models frequently concentrate on a small subset of valid outputs. In the professions task, some models repeatedly return the same individuals: for example, \llamafour names \textit{Stephen Hawking} in all $100$ completions for ``Write a short article about a well-known physicist'' (see Figure~\ref{fig:epistemic-examples}). Similarly, in the proofs dataset, models often produce only a single proof strategy (e.g., Euclid's proof), despite the existence of many valid alternatives~\citep{mestrovicEuclidsTheoremInfinitude2023}. This behavior reflects epistemic narrowness despite factual correctness.

\xhdr{Large portions of the answer space are never explored.}
Many valid answers are never produced by any model. In the professions dataset, notable individuals such as \textit{Hans Zimmer} and \textit{John Williams} are never mentioned, and across all models only $2$--$24\%$ of reference names are covered (Table~\ref{tab:wiki_coverage}). In the proofs dataset, several known proof strategies are similarly absent. This indicates that models fail to explore the long tail of valid answers.

\begin{figure*}[t]
    \centering

    \begin{subfigure}[t]{0.42\textwidth}
        \centering
        \includegraphics[width=\linewidth]{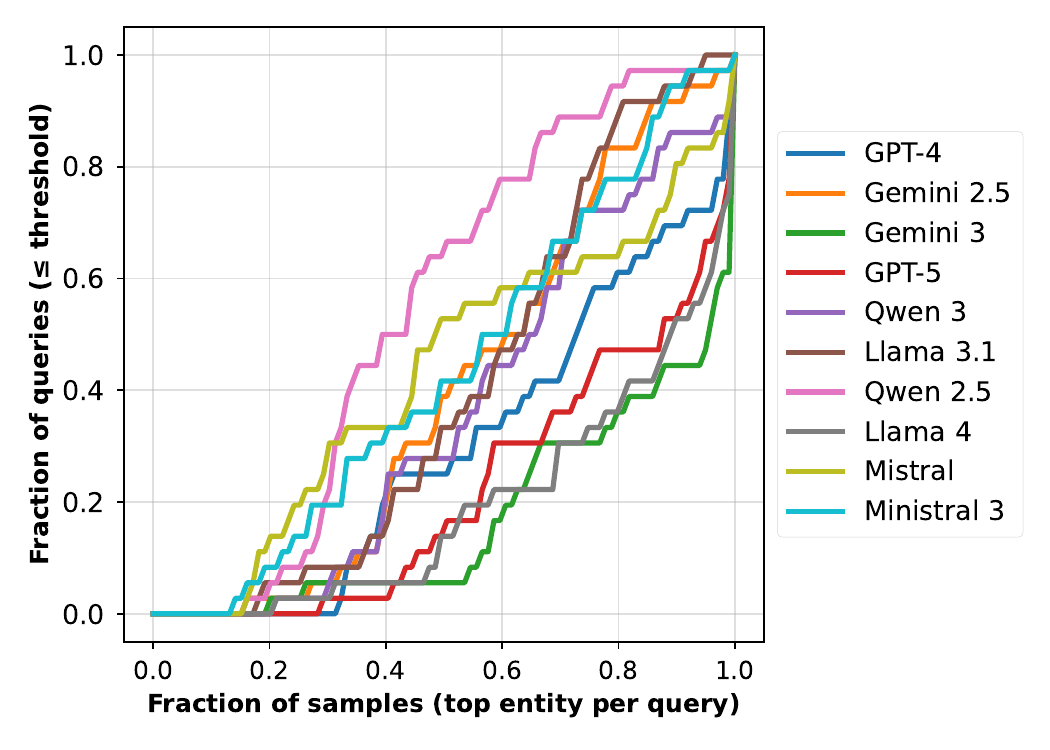}
        \caption{}
        \label{fig:entity-cdf}
    \end{subfigure}
    \hfill
    \begin{subfigure}[t]{0.56\textwidth}
        \centering
        \includegraphics[width=\linewidth]{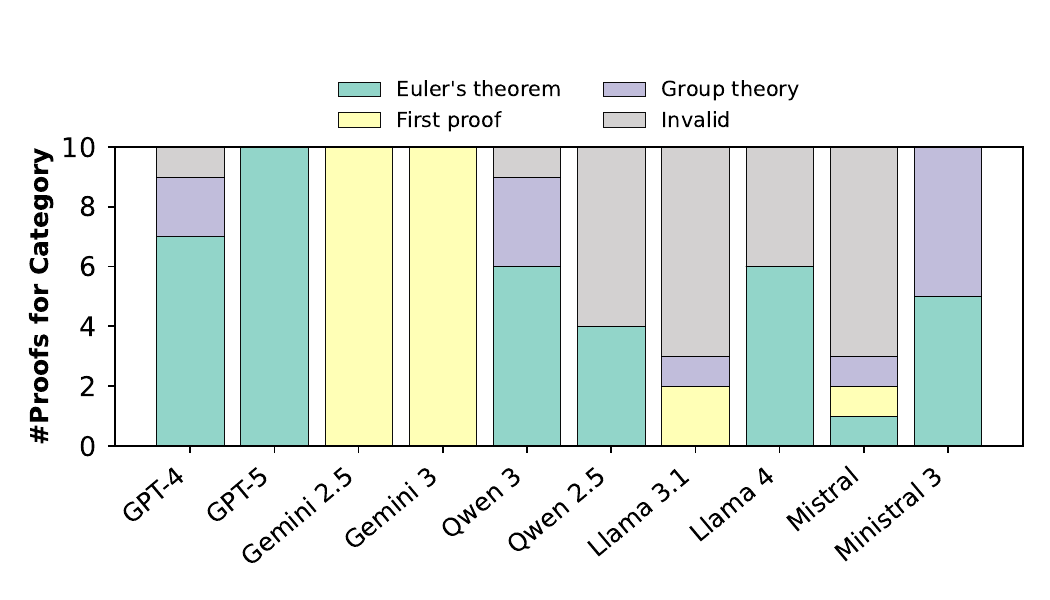}
        \caption{}
        \label{fig:stacked-proofs}
    \end{subfigure}
    \caption{
        Analysis of diversity and reasoning strategies across model responses.
        \textit{Left:} Maximum fraction of responses assigned to the most frequent entity for a given query (x-axis) \vs fraction of queries whose dominance is $\leq$ that level (y-axis) in the Professions dataset. Steeper curves indicate greater response diversity, flatter curves indicate stronger concentration on a single individual.
        \textit{Right:} Distinct proof categories used by each model when asked to solve Fermat's Little Theorem.
        GPT models prefer Euler's theorem, while Gemini takes an approach via induction (``first proof'').
         }
    \label{fig:combined-results}
\end{figure*}

\xhdr{Different models favor different regions of the answer space.}
Models exhibit systematic preferences over valid answers. For example, in the proofs dataset (Figure~\ref{fig:stacked-proofs}), \geminithree and \geminitwo consistently use induction-based proofs, while \gptfive and \gptfour prefer approaches via Euler's theorem. Users of different models may therefore be exposed to systematically different knowledge and reasoning strategies.

\begin{figure}
    \centering
    
    \begin{subfigure}{\linewidth}
        \centering
        \includegraphics[width=0.9\linewidth]{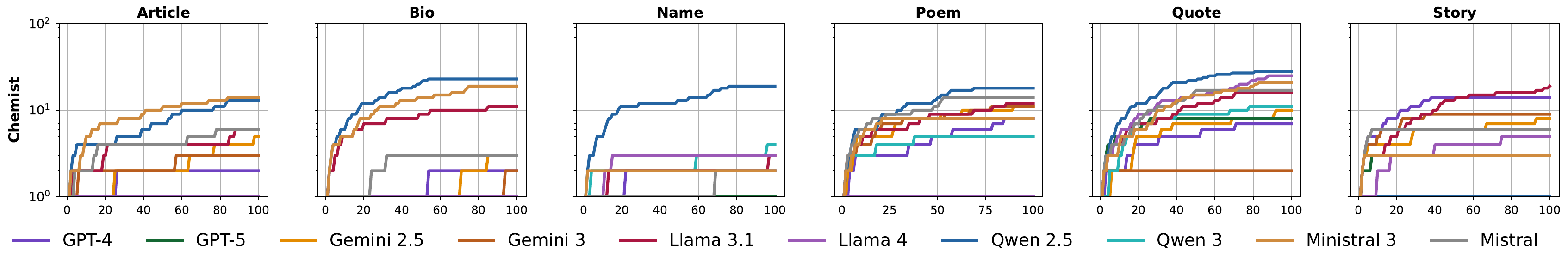}
        \caption{[\textbf{Accessible Diversity}] Cumulative gain over named valid entities across repeated generations. In most cases, marginal gain is sublinear.}
        \label{fig:marginal_gain_professions}
    \end{subfigure}
    
    \vspace{0.5em} %
    
    \begin{subfigure}{\linewidth}
        \centering
        \includegraphics[width=0.9\linewidth]{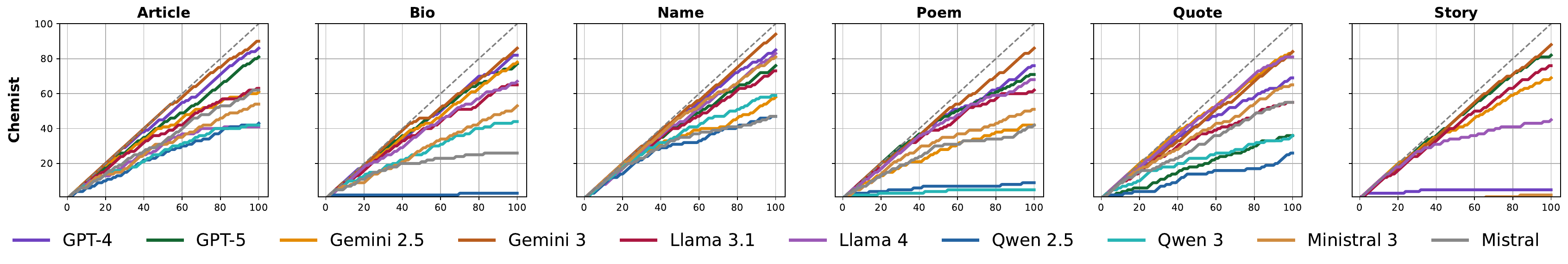}
        \caption{[\textbf{Latent Diversity}] Cumulative gain over named valid entities under iterative prompting. Models reach different numbers of names after 100 rounds.}
        \label{fig:latent_gains_row}
    \end{subfigure}
    
    \caption{Comparison of cumulative gains across interaction strategies. We report results for all professions in \Cref{fig:marginal_gain_professions_full,fig:latent_gains}.}
    \label{fig:combined}
\end{figure}

\begin{figure}
    \centering
    \includegraphics[width=0.5\linewidth]{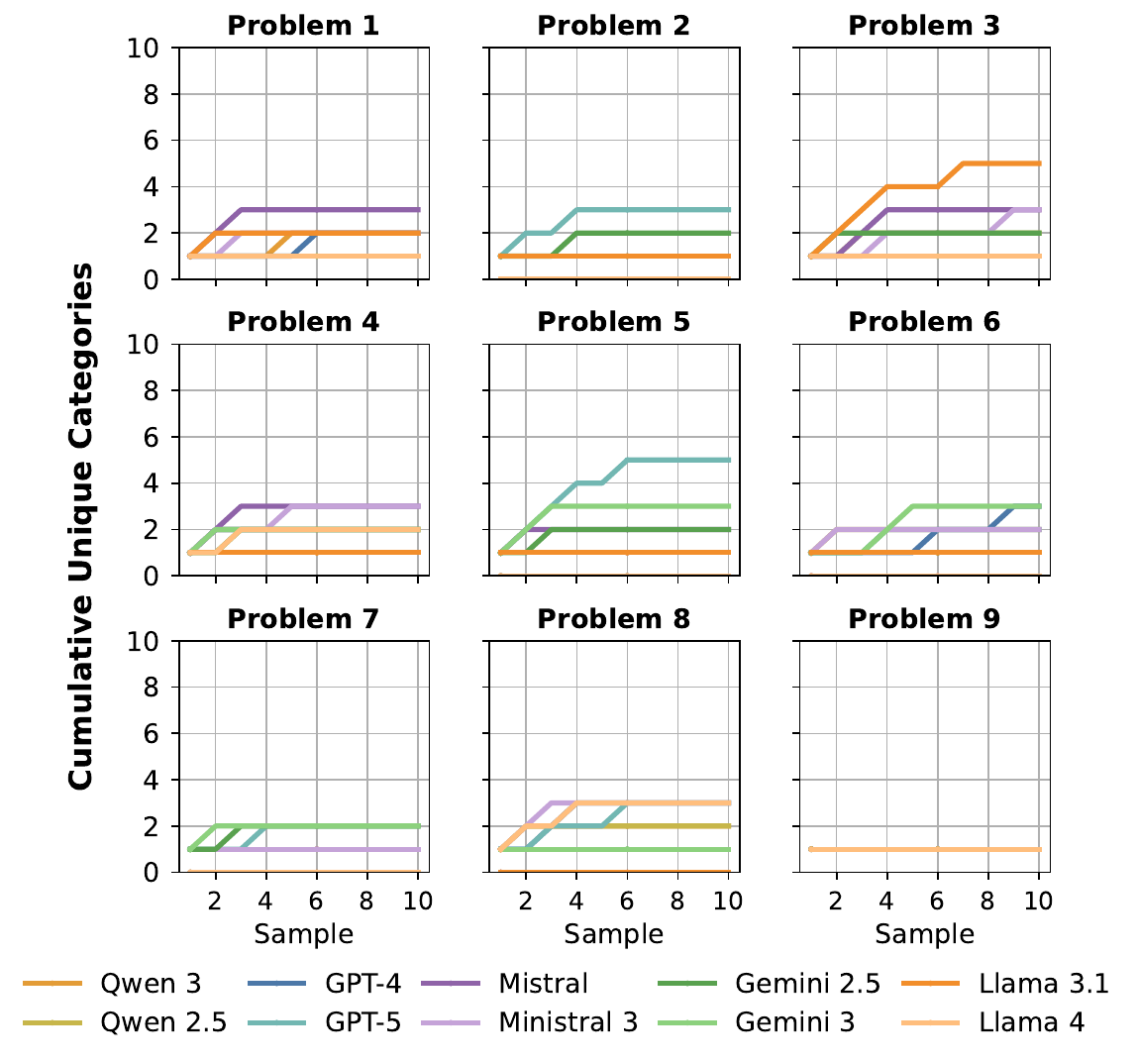}
    \caption{[\textbf{Accessible Diversity}] Cumulative count of unique proof categories across repeated generations.
    In most cases, models saturate early.}
    \label{fig:cum-proof}
\end{figure}

\begin{figure}[t]
    \centering

    \begin{minipage}[t]{0.33\linewidth}
        \vspace{0pt}
        \centering
        \includegraphics[
            width=\linewidth
        ]{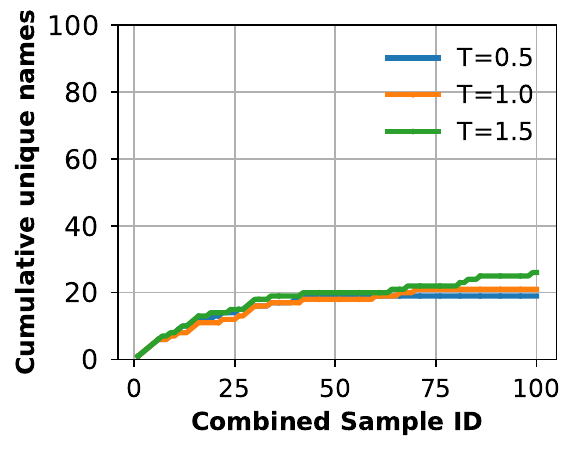}

        \captionof{figure}{
        Cumulative gains across samples for \geminithree{} at temperatures $\{0.5, 1.0, 1.5\}$. Varying the temperature has a limited effect on diversity.
        }
        \label{fig:temp-gemini}
    \end{minipage}
    \hfill
    \begin{minipage}[t]{0.60\linewidth}
        \vspace{0pt}
        \centering
        \scriptsize
        \setlength{\tabcolsep}{3pt}
        \renewcommand{\arraystretch}{1.05}

        \begin{tabularx}{\linewidth}{
            >{\raggedright\arraybackslash}X
            >{\centering\arraybackslash}p{0.55cm}
        }
            \toprule
            Text & $T$ \\
            \midrule

            \texttt{\#\# The Father of Modern Chemistry: Antoine Lavoisier}

            While many individuals have shaped the landscape of science, few have left a more profound mark on chemistry
            than Antoine Lavoisier'' [...]
            &
            $1.0$
            \\[1em]

            \texttt{\#\#\# **The Man Who Relabeled the World: Antoine Lavoisier**}

            In the late 18th century, chemistry was a tangled mess of ancient Greek theories and confusing ``mystical''   terminology. [...]
            &
            $1.5$
            \\

            \bottomrule
        \end{tabularx}

        \captionof{table}{
            Example generations for the query ``Write a story about a well-known chemist'' from \geminithree{} at different temperatures.
        }
        \label{tab:example_generations_temp}
    \end{minipage}
\end{figure}

\xhdr{Repeated sampling yields diminishing returns.}
Across both datasets, repeated sampling produces sublinear gains in diversity. In the professions task, the number of unique individuals quickly saturates with additional samples (\Cref{fig:marginal_gain_professions}). In the proofs dataset, most models converge to a small number of proof strategies (\Cref{fig:cum-proof}). Varying temperature has little effect on this behavior, indicating that stochasticity alone does not meaningfully expand the explored answer space (\Cref{fig:temp-gemini}).
Example outputs illustrate why: higher temperature changes the framing and prose of the generated text, but not the individual it is about (Table~\ref{tab:example_generations_temp}).

Figure~\ref{fig:entity-cdf} shows the maximum share of responses assigned to a single individual for each query: for a substantial fraction of queries, one individual accounts for the majority of all responses, with \llamafour and \geminithree the most concentrated.

These findings show that models tend to converge to narrow regions of the valid answer space, under-representing both entities and reasoning strategies.

\subsection{Accessible \vs Latent Diversity}

We next focus on the professions dataset and distinguish the diversity users see by default from the broader diversity models can reveal under stronger probing.

\begin{figure}
    \centering
    \includegraphics[width=.9\linewidth]{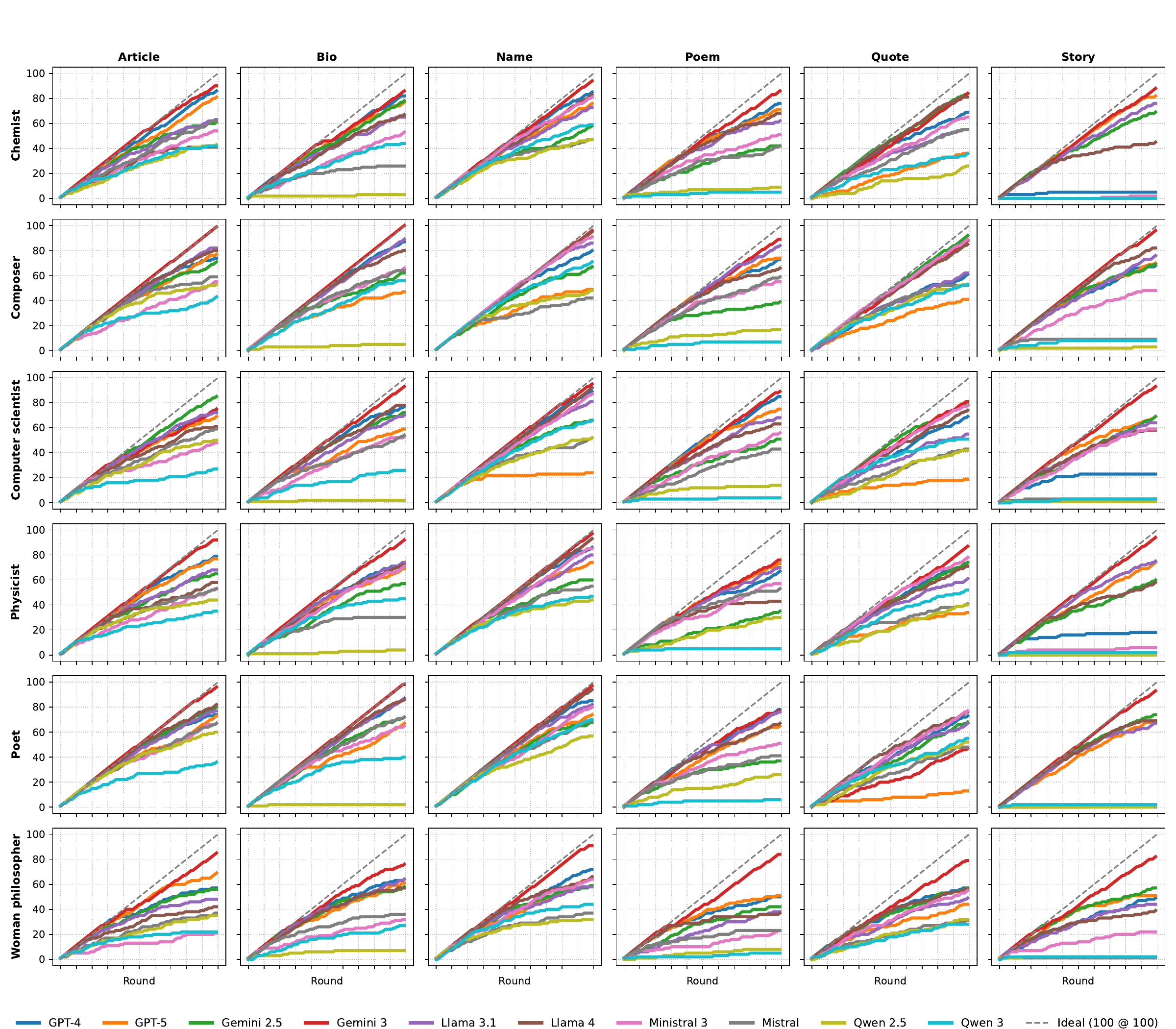}
    \caption{[\textbf{Latent Diversity}] Cumulative count of unique individuals under the iterative prompting protocol.}
    \label{fig:latent_gains}
\end{figure}

Accessible diversity is limited: even across repeated samples, models often return the same individuals, exposing users to only a narrow subset of valid answers.
Latent diversity is higher but still constrained. Under iterative prompting, models produce additional valid names but still fail to cover the full answer space (Figure~\ref{fig:latent_gains_row} and Figure~\ref{fig:latent_gains} for per-profession curves). 
For example, \geminithree produces on average 88 names, \llamathree 69, and \qwentwo and \qwenthree only 26 and 31,
often with hallucinated entries (Table~\ref{tab:unique_latent}).

We quantify this with the ratio between accessible and latent diversity (Table~\ref{tab:acc_latent_ratio}). The ratio is often low (e.g., $0.03$ for \geminithree), indicating that models expose only a small fraction of the diversity they can generate. 
A substantial accessible--latent gap persists overall.
Notably \qwentwo, which produced only a few valid names under the latent protocol, achieves a higher accessible--latent ratio.

\begin{table}[ht]
    \centering
    \resizebox{\textwidth}{!}{
    \begin{tabular}{llrrrrrrrrrr}
\toprule
 &  & \gptfour & \gptfive & \geminitwo & \geminithree & \llamathree & \llamafour & \ministral & \mistral & \qwentwo & \qwenthree \\
\midrule
\multirow[t]{6}{*}{article} 
 & chemist & 86 & 81 & 61 & \textbf{90} & 63 & 41 & 54 & 62 & 43 & 42 \\
 & composer & 74 & 77 & 71 & \textbf{99} & 82 & 80 & 55 & 59 & 53 & 43 \\
 & computer scientist & 74 & 69 & \textbf{85} & 75 & 72 & 61 & 48 & 59 & 50 & 27 \\
 & physicist & 79 & 77 & 65 & \textbf{92} & 68 & 58 & 53 & 53 & 44 & 35 \\
 & poet & 74 & 73 & 80 & \textbf{96} & 77 & 82 & 67 & 67 & 60 & 36 \\
 & woman philosopher & 57 & 69 & 56 & \textbf{85} & 48 & 42 & 21 & 37 & 35 & 22 \\
\cline{1-12}

\multirow[t]{6}{*}{bio} 
 & chemist & 82 & 77 & 78 & \textbf{86} & 65 & 67 & 53 & 26 & 3 & 44 \\
 & composer & 87 & 47 & 62 & \textbf{100} & 89 & 80 & 66 & 64 & 5 & 56 \\
 & computer scientist & 77 & 59 & 72 & \textbf{93} & 70 & 78 & 53 & 54 & 2 & 26 \\
 & physicist & 73 & 69 & 57 & \textbf{92} & 74 & 72 & 71 & 30 & 4 & 45 \\
 & poet & 87 & 67 & 72 & \textbf{98} & 87 & 86 & 65 & 72 & 2 & 40 \\
 & woman philosopher & 64 & 61 & 58 & \textbf{76} & 64 & 57 & 32 & 36 & 7 & 27 \\
\cline{1-12}

\multirow[t]{6}{*}{name} 
 & chemist & 85 & 76 & 58 & \textbf{94} & 73 & 83 & 81 & 47 & 47 & 59 \\
 & composer & 80 & 49 & 67 & \textbf{96} & 86 & 95 & 91 & 42 & 48 & 71 \\
 & computer scientist & 89 & 24 & 66 & \textbf{95} & 81 & 92 & 87 & 52 & 52 & 66 \\
 & physicist & 86 & 74 & 60 & \textbf{97} & 80 & 93 & 85 & 55 & 44 & 47 \\
 & poet & 85 & 74 & 68 & \textbf{97} & 82 & 94 & 80 & 69 & 57 & 70 \\
 & woman philosopher & 72 & 65 & 59 & \textbf{91} & 58 & 66 & 64 & 37 & 32 & 44 \\
\cline{1-12}

\multirow[t]{6}{*}{poem} 
 & chemist & 76 & 71 & 42 & \textbf{86} & 62 & 68 & 51 & 42 & 9 & 5 \\
 & composer & 73 & 74 & 39 & \textbf{89} & 84 & 66 & 55 & 59 & 17 & 7 \\
 & computer scientist & 85 & 75 & 51 & \textbf{89} & 68 & 63 & 56 & 43 & 14 & 4 \\
 & physicist & 67 & 73 & 35 & \textbf{76} & 70 & 43 & 57 & 53 & 30 & 5 \\
 & poet & 78 & 65 & 37 & \textbf{77} & 76 & 67 & 51 & 41 & 26 & 6 \\
 & woman philosopher & 50 & 51 & 42 & \textbf{84} & 38 & 37 & 23 & 23 & 8 & 5 \\
\cline{1-12}

\multirow[t]{6}{*}{quote} 
 & chemist & 69 & 36 & \textbf{84} & \textbf{84} & 55 & 81 & 65 & 55 & 26 & 36 \\
 & composer & 61 & 41 & \textbf{92} & 88 & 62 & 85 & 89 & 52 & 53 & 53 \\
 & computer scientist & 69 & 19 & 79 & \textbf{81} & 55 & 74 & 78 & 43 & 42 & 51 \\
 & physicist & 71 & 34 & 74 & \textbf{87} & 61 & 71 & 78 & 41 & 40 & 52 \\
 & poet & 73 & 13 & 68 & 47 & 67 & 76 & \textbf{77} & 48 & 52 & 55 \\
 & woman philosopher & 57 & 44 & 57 & \textbf{79} & 49 & 56 & 55 & 31 & 32 & 28 \\
\cline{1-12}

\multirow[t]{6}{*}{story} 
 & chemist & 5 & 82 & 69 & \textbf{88} & 76 & 45 & 2 & 0 & 0 & 0 \\
 & composer & 68 & 70 & 69 & \textbf{96} & 76 & 82 & 48 & 9 & 3 & 8 \\
 & computer scientist & 23 & 69 & 69 & \textbf{93} & 64 & 58 & 59 & 3 & 1 & 3 \\
 & physicist & 18 & 74 & 60 & \textbf{94} & 75 & 58 & 6 & 1 & 0 & 2 \\
 & poet & 1 & 69 & 74 & \textbf{93} & 67 & 69 & 2 & 0 & 1 & 2 \\
 & woman philosopher & 49 & 51 & 57 & \textbf{82} & 44 & 39 & 22 & 1 & 2 & 2 \\
\cline{1-12}

\bottomrule
\end{tabular}
}
    \caption{[\textbf{Latent Diversity}] Unique valid individuals generated by different models in the iterative prompting mode after 100 rounds (highest per row in bold). While \geminithree achieves the highest performance, other models fail to cover much of the valid answer space (\eg \qwentwo, \qwenthree).}
    \label{tab:unique_latent}
\end{table}

\begin{table}[ht]
    \centering
    \resizebox{\textwidth}{!}{
    \begin{tabular}{llrrrrrrrrrr}
\toprule
 &  & \gptfour & \gptfive & \geminitwo & \geminithree & \llamathree & \llamafour & \ministral & \mistral & \qwentwo & \qwenthree \\
\midrule
\multirow[t]{6}{*}{article}
 & chemist & 0.02 & 0.01 & 0.08 & 0.03 & 0.10 & 0.02 & 0.26 & 0.10 & \textbf{0.30} & 0.02 \\
 & composer & 0.03 & 0.03 & 0.08 & 0.02 & 0.13 & 0.02 & \textbf{0.16} & 0.10 & 0.15 & 0.07 \\
 & computer scientist & 0.07 & 0.03 & 0.14 & 0.08 & 0.08 & 0.05 & 0.19 & 0.05 & \textbf{0.30} & 0.15 \\
 & physicist & 0.04 & 0.03 & \textbf{0.18} & 0.03 & 0.06 & 0.02 & 0.09 & 0.13 & 0.11 & 0.17 \\
 & poet & 0.11 & 0.03 & 0.20 & 0.03 & 0.18 & 0.02 & \textbf{0.30} & 0.15 & 0.25 & 0.14 \\
 & woman philosopher & 0.07 & 0.04 & 0.23 & 0.04 & 0.12 & 0.02 & \textbf{0.48} & 0.22 & 0.43 & 0.41 \\
\cline{1-12}

\multirow[t]{6}{*}{bio}
 & chemist & 0.02 & 0.01 & 0.04 & 0.02 & 0.17 & 0.01 & 0.36 & 0.12 & \textbf{7.67} & 0.02 \\
 & composer & 0.02 & 0.06 & 0.06 & 0.02 & 0.12 & 0.02 & 0.21 & 0.16 & \textbf{4.40} & 0.04 \\
 & computer scientist & 0.13 & 0.03 & 0.10 & 0.02 & 0.16 & 0.03 & 0.36 & 0.09 & \textbf{13.50} & 0.23 \\
 & physicist & 0.08 & 0.03 & 0.07 & 0.01 & 0.09 & 0.03 & 0.14 & 0.17 & \textbf{3.00} & 0.11 \\
 & poet & 0.09 & 0.01 & 0.17 & 0.02 & 0.15 & 0.03 & 0.37 & 0.22 & \textbf{14.50} & 0.18 \\
 & woman philosopher & 0.03 & 0.03 & 0.10 & 0.01 & 0.08 & 0.04 & 0.50 & 0.11 & \textbf{2.86} & 0.37 \\
\cline{1-12}

\multirow[t]{6}{*}{name}
 & chemist & 0.02 & 0.01 & 0.02 & 0.01 & 0.04 & 0.04 & 0.02 & 0.04 & \textbf{0.40} & 0.07 \\
 & composer & 0.02 & 0.04 & 0.03 & 0.01 & 0.06 & 0.02 & 0.03 & \textbf{0.12} & 0.10 & 0.04 \\
 & computer scientist & 0.04 & 0.04 & 0.12 & 0.01 & 0.12 & 0.02 & 0.13 & 0.04 & \textbf{0.21} & 0.11 \\
 & physicist & 0.01 & 0.01 & 0.03 & 0.01 & 0.06 & 0.03 & 0.02 & 0.02 & \textbf{0.09} & 0.02 \\
 & poet & 0.05 & 0.01 & 0.06 & 0.03 & 0.11 & 0.05 & 0.09 & 0.10 & \textbf{0.26} & 0.06 \\
 & woman philosopher & 0.06 & 0.03 & 0.08 & 0.01 & 0.10 & 0.03 & 0.08 & 0.05 & 0.28 & \textbf{0.32} \\
\cline{1-12}

\multirow[t]{6}{*}{poem}
 & chemist & 0.11 & 0.01 & 0.26 & 0.13 & 0.19 & 0.01 & 0.16 & 0.33 & \textbf{2.00} & 1.00 \\
 & composer & 0.03 & 0.04 & 0.10 & 0.03 & 0.07 & 0.03 & 0.13 & 0.14 & 0.65 & \textbf{0.86} \\
 & computer scientist & 0.04 & 0.04 & 0.25 & 0.04 & 0.04 & 0.03 & 0.27 & 0.19 & \textbf{0.79} & 0.75 \\
 & physicist & 0.09 & 0.04 & 0.11 & 0.01 & 0.04 & 0.07 & 0.09 & 0.09 & 0.33 & \textbf{0.60} \\
 & poet & 0.09 & 0.12 & 0.41 & 0.05 & 0.24 & 0.09 & 0.35 & 0.39 & 0.85 & \textbf{2.50} \\
 & woman philosopher & 0.10 & 0.10 & 0.33 & 0.01 & 0.24 & 0.03 & 0.22 & 0.61 & 1.50 & \textbf{1.60} \\
\cline{1-12}

\multirow[t]{6}{*}{quote}
 & chemist & 0.10 & 0.22 & 0.12 & 0.02 & 0.29 & 0.31 & 0.32 & 0.31 & \textbf{1.08} & 0.31 \\
 & composer & 0.16 & 0.29 & 0.13 & 0.06 & \textbf{0.40} & 0.07 & 0.10 & 0.33 & 0.28 & 0.23 \\
 & computer scientist & 0.20 & 0.47 & 0.24 & 0.01 & 0.42 & 0.18 & 0.15 & 0.58 & \textbf{0.60} & 0.29 \\
 & physicist & 0.04 & 0.15 & 0.09 & 0.03 & \textbf{0.28} & 0.03 & 0.06 & 0.15 & 0.18 & 0.04 \\
 & poet & 0.12 & 0.38 & 0.28 & 0.02 & 0.18 & 0.04 & 0.25 & \textbf{0.62} & 0.31 & 0.45 \\
 & woman philosopher & 0.11 & 0.27 & 0.21 & 0.04 & 0.29 & 0.07 & 0.13 & 0.35 & \textbf{0.44} & 0.21 \\
\cline{1-12}

\multirow[t]{6}{*}{story}
 & chemist & \textbf{2.80} & 0.04 & 0.12 & 0.10 & 0.25 & 0.11 & 1.50 & -- & -- & -- \\
 & composer & 0.04 & 0.11 & 0.19 & 0.01 & 0.24 & 0.07 & 0.50 & 2.67 & \textbf{9.67} & 1.88 \\
 & computer scientist & 0.39 & 0.06 & 0.16 & 0.01 & 0.17 & 0.09 & 0.10 & 2.00 & \textbf{4.00} & 1.67 \\
 & physicist & 0.50 & 0.08 & 0.22 & 0.04 & 0.16 & 0.10 & 1.00 & \textbf{5.00} & -- & 2.50 \\
 & poet & \textbf{21.00} & 0.28 & 0.27 & 0.16 & 0.34 & 0.32 & 6.50 & -- & 12.00 & 3.00 \\
 & woman philosopher & 0.20 & 0.16 & 0.19 & 0.05 & 0.27 & 0.10 & 0.59 & \textbf{14.00} & 6.50 & 5.00 \\
\cline{1-12}

\bottomrule
\end{tabular}
}
    \caption{Accessible--latent ratio ($D_{acc}(Q)/D_{lat}(Q)$), with the best ratio per prompt in \textbf{bold}. ``--'' represents NaN values where latent diversity equals zero. In most cases, models exhibit significantly lower accessible diversity than they are capable of in iterative prompting.}
    \label{tab:acc_latent_ratio}
\end{table}

\begin{table}[t]
    \small
    \centering
    \begin{tabular}{lccc}
    \toprule
    Profession & Ref. List Size & Unique Named & \% Coverage \\
    \midrule
    Chemist & $670$ & $153$ & $23\%$ \\
    Computer Scientist & $708$ & $125$ & $18\%$ \\
    Physicist & $1077$ & $70$ & $6\%$ \\
    Composer & $4983$ & $121$ & $2\%$ \\
    Poet & $2499$ & $147$ & $6\%$ \\
    Woman Philosopher & $316$ & $77$ & $24\%$ \\
    \bottomrule
    \end{tabular}
    \caption{Coverage of Wikipedia reference lists by model completions (aggregated across all models and prompt formats).}
    \label{tab:wiki_coverage}
\end{table}

\begin{figure*}
    \centering
    \includegraphics[width=0.9\linewidth]{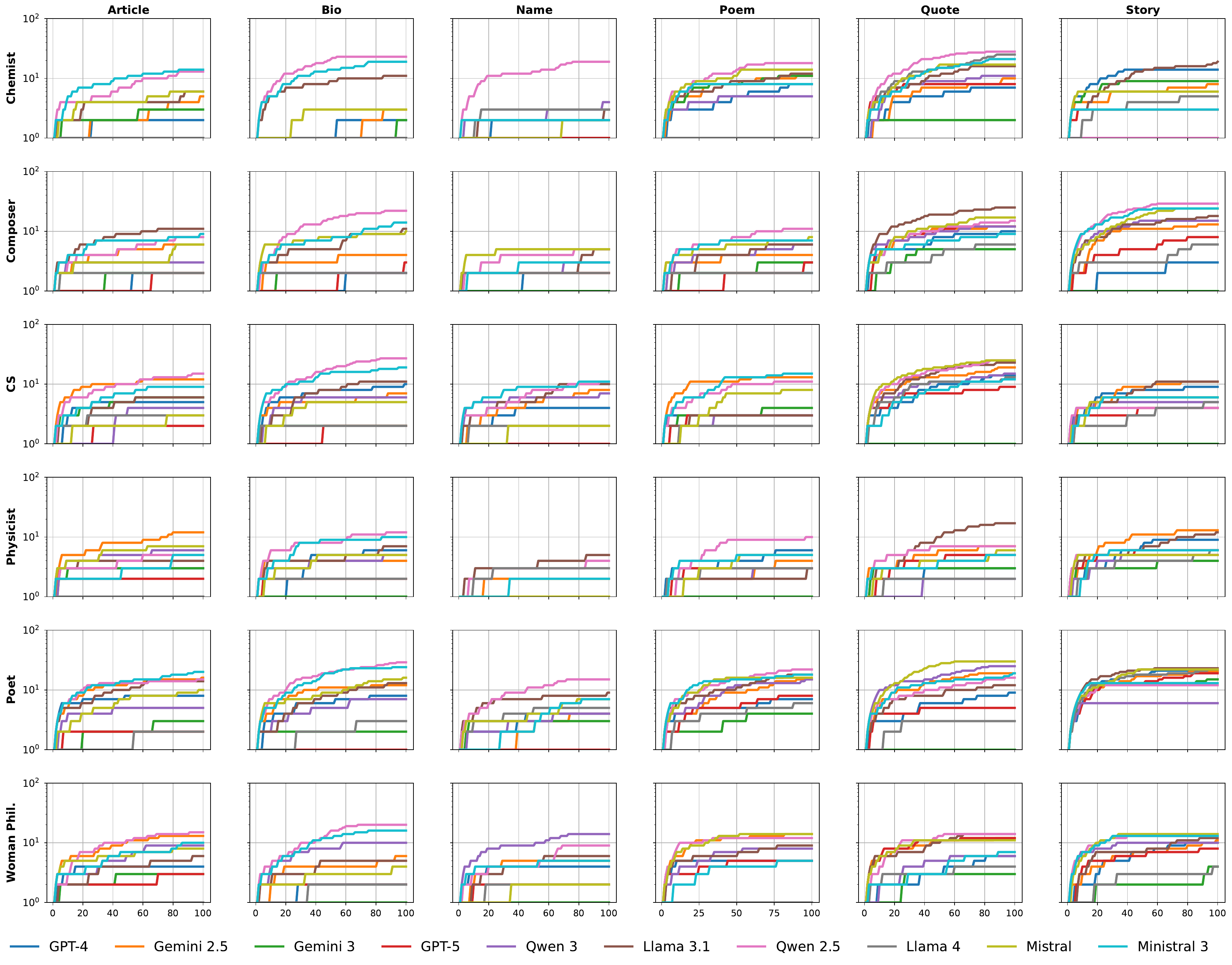}
    \caption{[\textbf{Accessible Diversity}] Cumulative count of unique individuals across repeated generations. In most cases, marginal gain is sublinear.}
    \label{fig:marginal_gain_professions_full}
\end{figure*}

\begin{figure}
    \centering
    \includegraphics[width=0.5\linewidth]{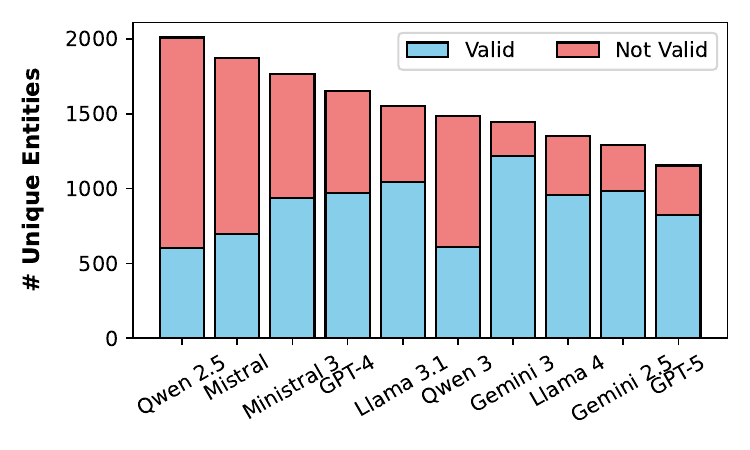}
    \caption{[\textbf{Latent Diversity}] Counts of valid and invalid unique names produced per model under iterative prompting.}
    \label{fig:latent_valid}
\end{figure}

\subsection{Failure Mode Analysis}
\label{app:failure-modes}

Figure~\ref{fig:latent_valid} shows the counts of invalid names produced per model after 100 rounds. Qwen models and \mistral show the highest invalid rates, with $71\%$ of names judged invalid for the profession for \qwentwo. In total, more than $90\%$ of names judged invalid are not present in our Wikipedia reference sets.
We provide a breakdown of what happens when the model no longer produces new valid names.
The most common failure modes we identify are repeated valid answers, repeated invalid answers and hallucinated persons, no extractable entities, and citing real persons outside of the target profession.

Across all models, $78\%$ of rounds successfully add a new name to the exclusion list, while $22\%$ stall, almost entirely because the model re-extracts someone already excluded ($17\%$) or the extractor returns no person ($5\%$). Stagnation rises sharply with round index and is worst for \qwenthree ($42\%$) and best for \geminithree ($5\%$), with duplicate repeats making up $\sim$ 80--97$\%$ of stalled rounds for strong models like \gptfour and \geminithree.

\begin{table}[t]
    \centering
    \begin{tabular}{lcccc}
    \toprule
         & Accessible & Latent & Multi-Turn & Multi-Output  \\ \midrule
    \gptfour        & 38 & \textbf{129} & 115 & 117 \\
    \gptfive        & 41 & \textbf{133} & 124 & 96 \\
    \geminitwo      & 62 & \textbf{150} & 148 & 100 \\
\geminithree        & 34 & 133 & \textbf{151} & 90\\
    \llamathree     & 62 & \textbf{144} & 139 & 120 \\
    \llamafour      & 30 & \textbf{134} & 113 & 101\\
    \ministral      & 68 & 141 & \textbf{156} & 104 \\
    \mistral        & 61 & \textbf{129} & 123 & 108 \\
    \qwentwo        & 85 & 106 & \textbf{134} & 126 \\
    \qwenthree      & 57 & 102 & \textbf{159} & 104 \\
    \bottomrule
    \end{tabular}
    \caption{Unique valid names per model after 10 rounds. Accessible refers to 10 independent samples, multi-turn to one conversation asking for alternatives, multi-output to asking for multiple outputs at once, and latent to iterative diversity-seeking prompting.}
    \label{tab:compare-protocols}
\end{table}

\begin{figure}[t]
    \centering
    \includegraphics[width=\linewidth]{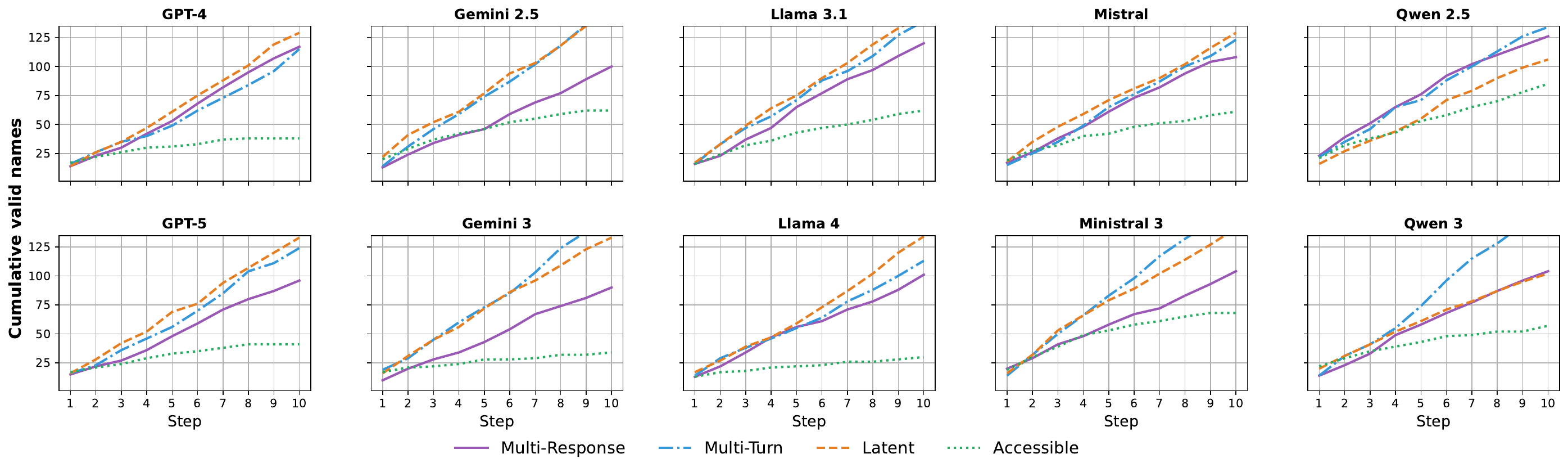}
    \caption{Cumulative number of distinct valid names as a function of step index (1–10), comparing multi-output, multi-turn, latent, and accessible protocols. For multi-output, each step adds the next sub-response from a single batched generation. Values are totals across all prompts and professions. }
    \label{fig:cumulative-four-strategies}
\end{figure}

\subsection{Diversity Under Explicit Exploration}
The latent protocol provides a controlled estimate of recoverable diversity, but it does not directly simulate ordinary user behavior. We therefore compare it to two additional diversity-seeking protocols on the Professions dataset: a multi-turn protocol, in which the user asks for alternatives in follow-up turns, and a multi-output protocol, in which the user requests several answers in a single completion.

At a matched budget of ten responses, explicit diversity-seeking substantially increases the number of unique valid names compared to accessible sampling. In the multi-turn protocol, models recover $8.2$ unique valid names on average, compared to $2.9$ under accessible sampling and $7.8$ under latent probing (averaged over the $36$ prompts and $10$ models). 
These results suggest that multi-turn interaction increases diversity compared to default prompting and recovers a similar amount of diversity compared to the latent protocol at the same budget.
Table~\ref{tab:compare-protocols} reports the number of recovered unique names per model for the accessible, latent, and multi-turn settings.
For six out of ten models, the latent protocol recovers more unique valid names in total after ten rounds. Notably, the multi-turn protocol reduces the number of invalid names produced by Qwen models.
Figure~\ref{fig:cumulative-four-strategies} shows the distinct valid canonical names accumulated over the first 10 responses, aggregated across prompts, for all models.

The multi-output protocol also increases diversity relative to accessible prompting. Its position relative to the other protocols depends on the level of aggregation.
Per model, it recovers fewer unique names than either latent or multi-turn probing (Table~\ref{tab:compare-protocols}).
Pooled across models, however, it yields more distinct valid names in total ($345$) than latent probing ($275$), and fewer than multi-turn prompting ($511$), against $227$ under accessible sampling.
Exclusion-based latent probing drives different models along the same long tail of the reference set, so their outputs overlap heavily, whereas the other protocols produce more model-specific sets.

Accessible prompting is clearly the most concentrated, with top-1 share $\sim11.5\%$ versus $\sim3$--$4\%$ top-1 share for multi-output, multi-turn, and latent protocols. 
These results show that epistemic narrowness is not fixed solely by the model's internal capabilities: the interaction protocol substantially shapes which parts of the valid answer space are surfaced.
At the same time, explicit exploration does not eliminate concentration entirely. Even when users ask for alternatives, models continue to favor a subset of highly salient individuals, and some parts of the reference space remain uncovered. Thus, the accessible--recoverable gap reflects both interface effects and persistent model-level priors over the answer space.

\subsection{Statistical Analysis}
\xhdr{Accessible diversity.}
For each model and prompt we compute the number of unique valid names recovered in ten samples, and
report the mean over the $36$ profession--format prompts with a bootstrap $95\%$ confidence interval
obtained by resampling prompts with replacement ($1000$ resamples; Table~\ref{tab:accessible-ci}).
These intervals reflect variation across prompts only. They do not capture uncertainty from the
choice of reference list or from the annotation judge, which we treat as fixed.

\xhdr{Comparing interaction protocols.}
Because every model is evaluated under every protocol, protocol comparisons are paired.
We therefore use the Wilcoxon signed-rank test with the model as the unit of analysis ($n = 10$),
which makes no distributional assumption and avoids treating the $360$ model--profession--format
cells as independent.
All three diversity-seeking protocols exceed accessible sampling for all ten models
($p = 0.002$, the smallest attainable value at $n = 10$). %
Multi-turn and latent probing show no detectable difference ($p = 0.70$; bootstrap $95\%$ CI on the
mean paired difference $[-6.7, +21.0]$ names), while the multi-output protocol recovers fewer unique
names per model than either.

\begin{table}[t]
    \centering
    \small
    \begin{tabular}{lcc}
        \toprule
        Model & Mean & CI \\
        \midrule
        \gptfour    & $2.50$ & $[2.08, 2.94]$ \\
        \gptfive    & $2.17$ & $[1.78, 2.61]$ \\
        \geminitwo  & $3.53$ & $[2.92, 4.19]$ \\
        \geminithree & $1.89$ & $[1.47, 2.39]$ \\
        \llamathree & $3.53$ & $[2.97, 4.14]$ \\
        \llamafour  & $1.97$ & $[1.56, 2.45]$ \\
        \ministral  & $3.22$ & $[2.69, 3.75]$ \\
        \mistral    & $2.92$ & $[2.33, 3.56]$ \\
        \qwentwo    & $3.97$ & $[3.36, 4.56]$ \\
        \qwenthree  & $2.78$ & $[2.22, 3.39]$ \\
        \bottomrule
    \end{tabular}
    \caption{Unique valid names recovered under the accessible protocol at a budget of ten samples,
    averaged over the $36$ profession--format prompts, with bootstrap $95\%$ confidence intervals
    over prompts.}
    \label{tab:accessible-ci}
\end{table}

\subsection{Human validation of automated annotation}
\label{app:human-eval}
To assess the reliability of our LLM-based annotation pipeline, two authors independently annotated $100$ randomly selected outputs: $50$ from the professions task and $50$ from the proofs task, with $5$ samples from each of the $10$ tested models per task. The annotators agreed in $98/100$ cases. The remaining two disagreements were resolved through discussion. The final labels agreed with the LLM judge annotations in $100/100$ cases.

\subsection{Impact of Phrasing Variations}
\label{app:phrasing-var}
We test whether small phrasing variations of the input prompts, such as ``a'' \vs ``any'' (\eg in ``Name a computer scientist'') affect diversity.
We only find small differences, yielding $32.7$ unique valid names on average (per model-protocol setting) for the ``any'' phrasing, compared to $33.2$ for the original ``a'' prompt (see Table~\ref{tab:a-any-diff}).

\begin{table*}[t]
    \centering
    \small
    \begin{tabular}{llrrrrr}
        \toprule
        \textbf{Model} &
        \textbf{Protocol} &
        \textbf{``a'' condition} &
        \textbf{``any'' condition} &
        \textbf{Overlap} &
        \textbf{Union} &
        \textbf{Jaccard} \\
        \midrule
        \gptfour     & latent@100      & 79 & 77 & 52 & 104 & 0.50 \\
        \gptfour     & multiturn@10    & 17 & 14 & 12 & 19  & 0.63 \\
        \gptfour     & accessible@100  & 4  & 6  & 4  & 6   & 0.67 \\
        \addlinespace
        \geminitwo & latent@100      & 72 & 75 & 51 & 96  & 0.53 \\
        \geminitwo & multiturn@10    & 22 & 21 & 13 & 30  & 0.43 \\
        \geminitwo & accessible@100  & 5  & 3  & 3  & 5   & 0.60 \\
        \bottomrule
    \end{tabular}
    \caption{Unique valid names recovered for the prompts ``Name a well-known 
    computer scientist'' and ``Name any well-known computer scientist'' across 
    two models. The latent and accessible protocols are executed for 100 rounds. 
    The multi-turn protocol consists of 10 conversations with 10 turns each.}
    \label{tab:a-any-diff}
\end{table*}

\end{document}